\documentclass[runningheads]{llncs}

\usepackage[final]{eccv}
\usepackage{eccvabbrv}
\usepackage{array}
\usepackage{booktabs}
\usepackage{makecell}
\usepackage{multirow}
\usepackage{graphicx}
\usepackage{tabularx}
\usepackage{pifont}
\usepackage{ragged2e}
\usepackage[section]{placeins}
\usepackage{needspace}
\usepackage{iftex}
\ifPDFTeX
  \usepackage[accsupp]{axessibility}
\else
  \usepackage{fontspec}
  \AtBeginDvi{}
\fi
\usepackage[final]{microtype}
\usepackage{wrapfig}
\usepackage[most]{tcolorbox}
\usepackage{listings}
\newcolumntype{M}[1]{>{\centering\arraybackslash}m{#1}}
\newcolumntype{P}[1]{>{\arraybackslash}m{#1}}

\usepackage[breaklinks,colorlinks,citecolor=eccvblue,linkcolor=eccvblue,urlcolor=eccvblue]{hyperref}

\lstdefinestyle{promptstyle}{
  basicstyle=\ttfamily\scriptsize,
  breaklines=true,
  breakatwhitespace=false,
  columns=fullflexible,
  keepspaces=true,
  showstringspaces=false,
  frame=none,
  upquote=true
}

\newtcolorbox{promptbox}[2][]{
  enhanced,
  breakable,
  colback=white,
  colframe=gray!80!black,
  colbacktitle=gray!80!black,
  coltitle=white,
  fonttitle=\bfseries,
  boxrule=0.8pt,
  arc=2.5mm,
  top=2mm,
  bottom=2mm,
  left=2mm,
  right=2mm,
  title={#2},
  title after break={#2\ \textit{(Continued)}},
  #1
}

\hypersetup{
  pdftitle={RVG-Bench: A Benchmark for Reasoning-Driven Audio-Visual Temporal Grounding},
  pdfauthor={Yue Feng, Boshen Xu, Zihao Yue, Xiaokun Liu, Miao Yang, Xin Tao, Pengfei Wan, Qin Jin},
  pdfsubject={Temporal video grounding},
  pdfkeywords={Temporal Video Grounding, Audio-Visual Understanding, Benchmark}
}

\begin{document}

\title{RVG-Bench: A Benchmark for Reasoning-Driven Audio-Visual Temporal Grounding}
\titlerunning{RVG-Bench}
\author{%
Yue Feng\inst{1} \and
Boshen Xu\inst{1} \and
Zihao Yue\inst{1} \and
Xiaokun Liu\inst{2*} \and\\
Miao Yang\inst{2} \and
Xin Tao\inst{2} \and
Pengfei Wan\inst{2} \and
Qin Jin\inst{1\dagger}}
\authorrunning{Y. Feng et al.}
\institute{AIM3 Lab, Renmin University of China \and Kling AI}

\maketitle
\begingroup
  \makeatletter
  \let\@footnotetext\H@@footnotetext
  \makeatother
  \renewcommand{\thefootnote}{\ensuremath{*}}
  \footnotetext[1]{Project leader.}
  \renewcommand{\thefootnote}{\ensuremath{\dagger}}
  \footnotetext[2]{Corresponding author.}
\endgroup

\begin{center}
\small\textbf{Project Page:} \url{https://fyrisk.github.io/RVG-Bench/}
\end{center}

\begin{abstract}
Existing temporal video grounding (TVG) benchmarks primarily focus on localizing target video segments according to descriptive text queries of the visual content.
However, real-world queries are more diverse and often require the model to first understand the user's intent before localization, such as finding visual evidence, identifying the cause of an event, or recalling a remembered moment, yet such queries remain largely overlooked.
To mitigate this gap, we introduce RVG-Bench, a benchmark for reasoning-driven audio-visual temporal grounding, where every query is constructed such that the target segment can only be identified through reasoning about the query's underlying intent.
We carefully construct the benchmark and rigorously verify its data quality to ensure models truly comprehend the audio-visual content.
RVG-Bench comprises 1,255 videos and 1,890 query-interval pairs across eight reasoning types.
We evaluate 22 models and find that the strongest, Qwen3.8-Omni-Flash, achieves only 49.6 mIoU. 
Through a series of controlled analyses, including oracle-grounding probes and test-time scaling experiments, we systematically examine where and why current models struggle, providing insights for future model development.

\keywords{Temporal Video Grounding \and Audio-Visual Understanding \and Benchmark}
\end{abstract}

\section{Introduction}
\label{sec:intro}

\begin{figure*}[t!]
  \centering
  \includegraphics[width=\textwidth,trim=20bp 13bp 38bp 8bp,clip]{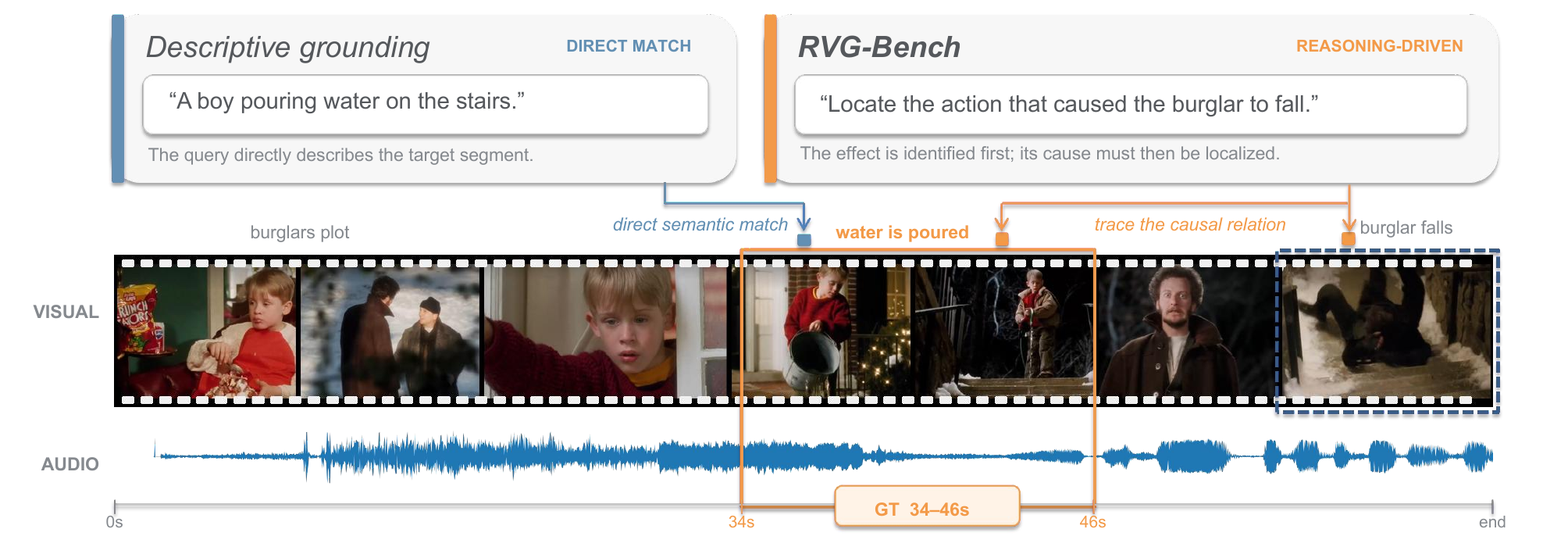}
    \vspace{-10pt}
  \caption{\textbf{From descriptive grounding to reasoning-driven grounding.} 
Left: Traditional TVG queries directly describe the target segment, making localization largely solvable via cross-modal semantic matching. 
Right: RVG-Bench uses indirect, intent-driven queries that require reasoning to infer the correct target interval, such as identifying the causal event that leads to a later outcome, and often rely on both visual and audio evidence.}
  \label{fig:figure1}
  \vspace{-7pt}
\end{figure*}

Temporal Video Grounding aims to localize a temporal segment in an untrimmed video given a natural language query. In existing benchmarks, the query explicitly describes the visual content of the target segment, such as ``A boy pouring water on the stairs,'' shown on the left of \cref{fig:figure1}. This formulation effectively reduces grounding to \emph{description-to-segment matching}, where the model aligns textual descriptions with video frames through cross-modal semantic similarity.

However, real-world video grounding rarely follows this pattern. Users typically do not describe what a segment looks like; instead, they express an \emph{intent}: locating evidence for a question, identifying the cause of an observed outcome, or recalling a moment from a vague memory. Such queries are inherently indirect. For example, ``Locate the action that caused the burglar to fall'' in \cref{fig:figure1} requires tracing the fall back to the boy pouring water on the stairs, then localizing that earlier action. This reasoning-driven grounding is a core capability required by many downstream video understanding tasks, yet it remains unevaluated by existing benchmarks.

Early temporal grounding models relied on task-specific architectures with limited generalization. The advent of Multimodal Large Language Models has fundamentally changed this landscape: models such as Gemini-3-Pro~\cite{googledeepmind2026gemini3pro} now possess strong video perception, cross-modal reasoning, and instruction-following capabilities within a single unified framework. Established benchmarks such as Charades-STA~\cite{sigurdsson2016hollywood} and ActivityNet Captions~\cite{krishna2017dense} primarily evaluate \emph{descriptive grounding}. Strong performance on these benchmarks provides limited insight into whether models can handle the intent-driven, reasoning-intensive queries that practical video understanding applications demand.

To address this gap, we propose \emph{Reasoning Video Grounding (RVG)}, a task in which the query does not directly describe the target segment but instead requires reasoning about the user's underlying intent to infer which event should be localized. Unlike descriptive grounding, where the target is explicitly depicted in the query, RVG requires models to jointly perform perception, reasoning, and temporal localization.
We argue that \emph{reasoning grounding} is a more fundamental capability and should be prioritized in both model training and evaluation.

To systematically evaluate this capability, we introduce RVG-Bench, a benchmark designed around a key principle: the target segment cannot be identified through surface-level semantic matching and must instead be inferred through reasoning. Every query is intentionally indirect, requiring the model to interpret the underlying intent before localization becomes possible.
To comprehensively cover the different dimensions of reasoning ability, we organize the benchmark using a structured reasoning taxonomy with three major categories: relational, intrinsic-attribute, and relative-attribute. These categories are further divided into eight subcategories. In addition, we incorporate the \emph{audio modality}, since narration, dialogue, and environmental sounds frequently provide critical cues that are absent from the visual stream. More than half of the queries therefore require audio evidence, either alone or jointly with vision. To ensure dataset reliability, we implement strict information-leakage prevention and multi-stage verification so that each query corresponds to a unique and unambiguous target segment. RVG-Bench contains 1,255 videos and 1,890 query--interval pairs constructed through a four-stage pipeline with full human validation.

We benchmark 22 models on RVG-Bench through APIs and local inference, covering both vision-language and omni-modal architectures. Reasoning grounding remains a substantial challenge: the strongest model, Qwen3.8-Omni-Flash, achieves 49.6 mIoU, while the strongest locally evaluated baseline reaches 30.1 mIoU. Beyond reporting overall performance, we conduct controlled analyses to examine target inference and temporal localization. Finally, we explore task decomposition at test time. These analyses provide concrete directions for improving reasoning grounding in future models.

In summary, our contributions are threefold: 
\begin{itemize}
    \item We introduce \emph{reasoning video grounding}, a new task that shifts temporal localization from description matching to intent-driven reasoning.
    \item We construct RVG-Bench, a benchmark of 1,255 videos and 1,890 query-interval pairs with a structured reasoning taxonomy, strict quality control against information leakage, and native audio-visual evaluation, providing a more realistic testbed for video understanding models.
    \item We conduct comprehensive evaluations and diagnostic analyses across a wide range of models,  revealing the current limitations of reasoning grounding and exploring test-time scaling strategies that offer directions for future model development.
\end{itemize}
We will publicly release RVG-Bench, the complete construction and evaluation prompts, and code for data construction and evaluation.

\section{Related Work}

Temporal Video Grounding (TVG) requires localizing specific segments in a video given textual queries. A large number of TVG datasets have been proposed, among which Charades-STA~\cite{sigurdsson2016hollywood}, ActivityNet Captions~\cite{krishna2017dense}, and QVHighlights~\cite{lei2021detecting} are the most widely adopted. However, these benchmarks are primarily vision-centric, with queries describing visually observable actions or events in silent videos. As multimodal models evolved, more recent efforts began incorporating audio signals for temporal grounding~\cite{geng2023dense,tian2018audio,geng2025longvale,chen2025chronusomni}, aiming to capture richer cross-modal cues.

Despite this progress, existing TVG benchmarks share a common limitation: their queries are predominantly descriptive, requiring only surface-level matching between the query text and video content, while reasoning-driven localization remains largely unexplored. Grounded Video QA datasets such as NEXT-GQA~\cite{xiao2024can} and REXTIME~\cite{chen2024rextime} attempt to bridge this gap by combining question answering with temporal localization. However, their multiple-choice format inevitably leaks answer cues, failing to reflect the open-ended nature of real-world queries. Similarly, while ToTG-Bench~\cite{liu2025timescope} explores related task-oriented localization, it lacks a systematic emphasis on the complex reasoning required to resolve each query. Furthermore, all of these aforementioned benchmarks are vision-centric, overlooking the audio modality. To address these limitations, we introduce RVG-Bench, which requires models to perform open-ended, reasoning-driven temporal localization with strong reliance on joint audio-visual evidence. A detailed comparison is provided in~\cref{tab:benchmark_comparison}.

\section{RVG-Bench}
\label{sec:rvg-bench}

\subsection{Benchmark Design}
\label{sec:benchmark-design}

\begin{table*}[tb]
\centering
\footnotesize
\caption{\textbf{Compact reasoning taxonomy of RVG-Bench.} Each subcategory is defined by how the query indirectly specifies the target. Complete definitions and example queries are provided in Appendix~\ref{sec:supp_full_taxonomy}.}
\label{tab:taxonomy_compact}
\setlength{\tabcolsep}{3pt}
\renewcommand{\arraystretch}{1.0}
\begin{tabularx}{\linewidth}{@{}P{1.75cm} M{1.4cm} >{\raggedright\arraybackslash}X@{}}
\toprule
\textbf{\makecell{Major\\Category}} & \textbf{\makecell{Sub-\\category}} & \textbf{Definition} \\ 
\midrule
\multirow{3}{=}{\makecell{Intrinsic\\Attribute}} 
& Char & Identify a segment through a character's implicit mental state or behavioral pattern. \\ \cmidrule{2-3}
& Anom & Locate a segment that deviates from the expected norm of surrounding context. \\ \cmidrule{2-3}
& Symb & Interpret abstract or metaphorical meaning to identify the target. \\
\midrule
\multirow{2}{=}{\makecell{Relative\\Attribute}}
& Comp & Select a segment by comparing a measurable property across candidates. \\ \cmidrule{2-3}
& Ord & Locate a segment by its position in a sequence. \\
\midrule
\multirow{3}{=}{\makecell{Relational}}
& Caus & Trace from an effect to its cause, or vice versa. \\ \cmidrule{2-3}
& Seq & Identify a segment by temporal adjacency to a specified event. \\ \cmidrule{2-3}
& Ref & Resolve an anaphoric reference to locate the referred segment. \\
\midrule
\makecell{Multi-step} & M-Step & Compose two or more atomic reasoning skills within a single query. \\
\bottomrule
\end{tabularx}
\end{table*}

In conventional descriptive grounding, the query directly specifies the target segment, typically through a caption that can be matched to visual content using surface-level similarity. In contrast, reasoning grounding requires indirect inference to identify the correct segment. We categorize such reasoning into three fundamental types based on how the query relates to its target.

\emph{Intrinsic-attribute reasoning.} The query refers to an implicit property of the target segment, such as a character's emotion, an anomalous pattern, or a symbolic meaning, rather than its observable content. The model must infer what the segment contains from what it \emph{implies}.

\emph{Relative-attribute reasoning.} The query requires comparing attributes across multiple candidate segments, such as identifying the highest jump or the last occurrence. The target is therefore defined by its relative ranking among alternatives rather than its standalone content.

\emph{Relational reasoning.} Events in a video form structured dependencies: causes precede effects, actions follow intentions, and references link to prior mentions. The query specifies one element in this structure and requires traversing a causal, temporal, or referential relation to locate the target.

These three categories reflect increasing complexity in how the query relates to its target: from reasoning over implicit properties of a single segment, to comparing across segments, to navigating dependencies among segments. In RVG-Bench, we view each of these fundamental operations as an atomic \textbf{\emph{single-step reasoning}} skill, where the target can be identified via a single inference. However, real-world user intents often require richer compositionality. To better capture this complexity, our benchmark further introduces \textbf{\emph{multi-step reasoning}}, in which a query demands chaining two or more atomic operations to uniquely identify the target segment. For example, the model may first infer a character's emotion and then locate the event that caused it. Each category is further divided into fine-grained subcategories detailed in \cref{tab:taxonomy_compact}.

\subsection{Benchmark Construction}
\label{sec:benchmark-construction}
\begin{figure}[tb]
  \centering
  \includegraphics[width=\linewidth]{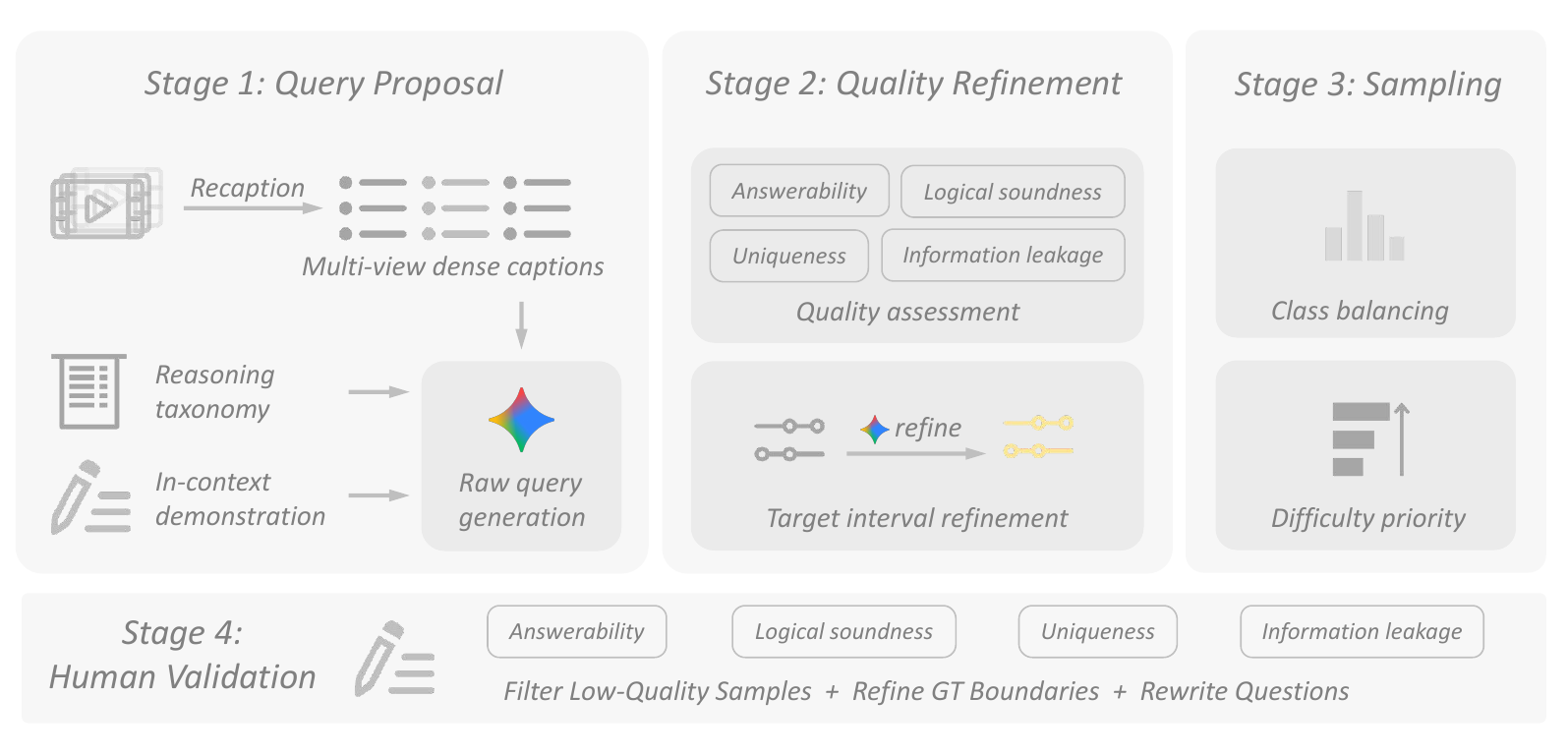}
  \vspace{-2mm}
  \caption{Overview of the RVG-Bench construction pipeline: (1) broad query proposal via multi-view captions; (2) rigorous quality filtering and target interval refinement using frontier MLLMs; (3) difficulty-aware sampling; and (4) full human validation with low-quality sample removal, question rewriting, and GT boundary refinement.}
  \label{fig:pipeline}
  \vspace{-3mm}
\end{figure}
As illustrated in \cref{fig:pipeline}, we construct RVG-Bench via a four-stage pipeline powered by frontier MLLMs and full human validation, using Gemini-2.5-Pro for proposal generation and Gemini-3-Pro for verification and refinement. Since reasoning-grounding queries are inherently indirect and cannot be trivially crowd-sourced, we leverage the strong narrative understanding and inference capabilities of frontier models to synthesize high-quality data at scale. To ensure rich narrative structures and sufficient audio evidence, candidate videos are sourced from established audio-visual QA and video reasoning benchmarks, including WorldSense~\cite{hong2025worldsense}, Daily-Omni~\cite{zhou2025daily}, FunQA~\cite{xie2024funqa}, and Video-Holmes~\cite{cheng2025video}. After filtering videos that are excessively short or long, we obtain an initial pool of 3,105 candidates. To ensure unique localizability, valid reasoning paths, and the absence of surface-level shortcuts, we combine broad proposal generation, rigorous quality refinement, difficulty-aware sampling, and full human validation. Videos that fail to produce valid reasoning instances are naturally discarded, yielding a final set of 1,255 videos.

\noindent\textbf{Stage~1: Query Proposal.}
To fully exploit the model's narrative understanding and reasoning capabilities in the text domain, we first convert each video into timestamped dense captions. Specifically, Gemini-2.5-Pro is prompted to independently generate three types of captions: visual events, audio events, and joint audio--visual events. While dense captioning may lose some fine-grained temporal precision, it preserves the narrative structure of the video and substantially reduces the cognitive burden on the model compared to raw video input.


Using these captions as context, Gemini-2.5-Pro generates candidate query-interval pairs. Because different videos support different reasoning types, we provide the model with definitions and demonstrations of all subcategories and allow it to adaptively select suitable query types based on the video content. If the context lacks sufficient information to support meaningful reasoning, the model abstains from generation and the video is discarded.

Generating multi-step reasoning queries is particularly challenging. Direct prompting often fails to produce genuine multi-hop reasoning. 
To address this issue, we introduce an intermediate \textit{event graph} representation. The model first parses dense captions into a structured graph where nodes represent events and their implicit attributes, such as emotions and symbolic meanings, and edges represent inter-event relations such as causal, temporal, or comparative dependencies. Multi-step queries are then constructed by composing multiple nodes and edges. For example, the model may first infer an implicit attribute and then traverse a causal relation. 
This proposal stage produces approximately 6.5k single-step and 1.1k multi-step query candidates, each paired with a draft temporal interval. Although the benchmark evaluation only requires temporal localization, we additionally ask the model to generate a textual answer describing the reasoning trace and the content of the target segment, which we later use for analysis.

\noindent\textbf{Stage~2: Quality Refinement.}
We apply two complementary refinement procedures using the more capable Gemini-3-Pro.

\noindent\textit{Quality filtering.} We enforce four criteria to ensure data quality: (i)~\emph{Answerability}: the query must be answerable from the video, with a correct answer corresponding to a localizable segment; (ii)~\emph{Uniqueness}: exactly one segment in the video satisfies the query; (iii)~\emph{Logical soundness}: the reasoning path must be valid, aligned with the designated category, and consistent with the audio-visual evidence; and (iv)~\emph{No information leakage}: the query must not directly reveal the visual or audio content of the target segment. For each criterion, we carefully design dedicated verification prompts and instruct \allowbreak Gemini-3-Pro to judge candidate samples against the dense captions. After this refinement stage, 66.3\% of the proposed samples are retained.

\noindent\textit{Target interval refinement.} The draft intervals produced in Stage~1 are derived from dense-caption timestamps and may not precisely correspond to atomic events. We therefore provide the raw video to Gemini-3-Pro and request refined start and end timestamps.

\noindent\textbf{Stage~3: Difficulty-Aware Sampling.}
From the refined pool, we perform category-balanced sampling to construct a candidate benchmark. Within each category, we prioritize more challenging instances using category-specific difficulty criteria. For example, in causal reasoning, we favor samples where the cause and effect segments are separated by a larger temporal gap. In character-centric reasoning, we prefer queries that require interpreting subtle micro-expressions or tone of voice rather than explicit dialogue. This stage produces 2,500 candidate query-interval pairs for human review.

\noindent\textbf{Stage~4: Human Validation.}
We conduct full human validation over all 2,500 sampled candidates using four criteria: Answerability, Uniqueness, Logical soundness, and No information leakage. Annotators remove samples that fail any criterion, rewrite questions that are valid but contain wording issues, and manually refine GT temporal boundaries. In total, 610 candidates, accounting for 24.40\%, are removed, yielding 1,890 final query-interval pairs, including 1,588 single-step and 302 multi-step queries. Among the retained samples, GT boundaries for 240 samples, accounting for 12.70\%, were refined, and questions for 98 samples, accounting for 5.19\%, were rewritten. Detailed validation statistics and the annotation interface are provided in Appendix~\ref{sec:supp_human_validation}.

\subsection{Statistics}
\label{sec:statistics}
RVG-Bench comprises 1,255 untrimmed videos paired with 1,890 reasoning query-interval annotations, including 1,588 single-step and 302 multi-step queries. A statistical overview is provided in \cref{fig:benchmark_statistics} in the appendix. As shown in \cref{fig:stats-vlen}, the videos have an average duration of 108.39 seconds, while the ground-truth intervals average only 9.46 seconds, as shown in \cref{fig:stats-gtlen}. Averaged over query--interval pairs, the intervals occupy approximately 9.6\% of their videos' duration. This large temporal gap requires models to perform precise localization over long contexts rather than relying on coarse segment matching. Moreover, ground-truth segments are approximately uniformly distributed across the timeline, reducing the usefulness of simple positional priors such as favoring the beginning or end of a video. The ground-truth temporal position distribution is shown in \cref{fig:gt_position_dist} in the appendix.

Regarding modality requirements, we use Gemini-3-Pro to annotate each query with the modality necessary for correct localization: visual-only, audio-only, or audio-visual joint. In RVG-Bench, 45.34\% of queries are visual-only, 11.38\% are audio-only, and 43.28\% require joint audio-visual reasoning. Thus, more than half of the queries require audio evidence, either alone or jointly with vision, confirming that RVG-Bench requires genuine omni-modal understanding rather than reliance on a single modality. The per-category modality distribution is reported in \cref{fig:stats-catmod}, and we further analyze the impact of audio information through modality ablation experiments in \cref{sec:benchmark-analysis}.

\section{Experiment}
\subsection{Setup}

\noindent\textbf{Models.} 
We evaluate 22 MLLMs in three groups. \textbf{API-based models} include Gemini-3-Pro~\cite{googledeepmind2026gemini3pro}, GPT-5.5~\cite{openai2026gpt55}, Qwen3.8-Omni-Flash~\cite{alibaba2026qwen38omni}, Qwen3.8-Flash~\cite{alibaba2026qwen38flash}, Seed2.0-Lite~\cite{bytedance2026seed20}, and Seed2.0-Pro~\cite{bytedance2026seed20}. \textbf{Locally evaluated omni models} include ChronusOmni~\cite{chen2025chronusomni}, Qwen3-Omni~\cite{xu2025qwen3}, Qwen2.5-Omni~\cite{xu2025qwen2}, HumanOmniV2~\cite{yang2025humanomniv2}, Ola-7B~\cite{liu2025ola}, and MiniCPM-O-2.6~\cite{yao2024minicpm}. \textbf{Locally evaluated vision-language (VL) models} include TimeLens-8B/7B~\cite{zhang2025timelens}, Qwen3-VL-8B~\cite{bai2025qwen3}, Time-R1-7B~\cite{wangtime}, TRACE~\cite{guo2024trace}, Qwen2.5-VL-7B~\cite{bai2025qwen2}, VideoChat-R1~\cite{li2025videochat}, TimeSuite~\cite{zeng2024timesuite}, Video-R1-7B~\cite{feng2025video}, and MiMo-VL-7B-RL~\cite{xiaomi2025mimovl}.

\noindent\textbf{Metrics and Experiment Settings.}
Following standard temporal grounding protocols~\cite{ren2024timechat,zeng2024timesuite,wangtime}, we report mean Intersection over Union (mIoU) and Recall at IoU@0.3, measuring the temporal alignment between predicted segments and ground-truth annotations.

Locally evaluated models use their official codebases and released checkpoints with default settings. API models receive visual or audio-visual inputs as indicated in \cref{tab:main_miou}; model inputs, sampling settings, and scoring protocols are detailed in Appendix~\ref{sec:supp_impl_details}. Unless otherwise specified, we use Chain-of-Thought (CoT) prompting, asking models to reason before producing timestamps. Prompt templates and results for alternative prompting strategies are provided in Appendices~\ref{sec:supp_eval_prompts}, \ref{sec:supp_full_tts}, and~\ref{sec:detailed_test_time_scaling}.


\begin{table*}[t!]
\centering
\caption{\textbf{Main results on RVG-Bench.} \textbf{Input} denotes visual (V) or audio-visual (AV) input. \textbf{S-Step} and \textbf{M-Step} denote single- and multi-step reasoning. Except for \textbf{IoU@0.3}, all numbers are mIoU percentages. Overall scores are sample-count-weighted. Additional recall results appear in Table~\ref{tab:strict_recall}.}
  \vspace{-8pt}
\label{tab:main_miou}

\footnotesize
\setlength{\tabcolsep}{2.6pt}
\renewcommand{\arraystretch}{1.0}

\resizebox{\textwidth}{!}{%
\begin{tabular}{lcrrrrrrrrrrrr}
\toprule
\multirow{2}{*}{\textbf{Model}} &
\multirow{2}{*}{\textbf{Input}} &
\multicolumn{9}{c}{\textbf{S-Step}} &
\multicolumn{1}{c}{\textbf{M-Step}} &
\multicolumn{2}{c}{\textbf{Overall}} \\
\cmidrule(lr){3-11}\cmidrule(lr){12-12}\cmidrule(lr){13-14}
& & \textbf{Caus} & \textbf{Seq} & \textbf{Ref} & \textbf{Char} & \textbf{Anom} & \textbf{Symb} & \textbf{Comp} & \textbf{Ord} & \textbf{Overall}
& \textbf{mIoU}
& \textbf{mIoU} & \textbf{IoU@0.3} \\
\midrule

\multicolumn{14}{c}{\textbf{API-based models}} \\
\midrule
Qwen3.8-Omni-Flash & AV & 38.5 & 71.1 & 47.1 & 41.9 & 43.8 & 58.6 & 62.9 & 51.9 & 51.3 & 40.6 & 49.6 & 68.0 \\
Seed2.0-Lite & AV & 36.8 & 64.2 & 50.2 & 40.0 & 44.6 & 56.2 & 66.7 & 42.7 & 49.0 & 36.9 & 47.0 & 62.6 \\
Gemini-3-Pro & AV & 31.3 & 63.5 & 44.4 & 34.6 & 41.4 & 51.0 & 68.3 & 45.1 & 46.1 & 35.8 & 44.4 & 61.1 \\
Seed2.0-Pro & V & 28.9 & 57.9 & 42.3 & 36.4 & 40.8 & 48.8 & 58.6 & 36.8 & 42.6 & 38.0 & 41.9 & 58.2 \\
GPT-5.5 & V & 28.2 & 50.9 & 38.0 & 32.8 & 43.4 & 47.9 & 50.1 & 32.6 & 39.5 & 32.8 & 38.5 & 55.5 \\
Qwen3.8-Flash & V & 26.5 & 48.1 & 40.6 & 33.4 & 31.5 & 49.2 & 57.9 & 33.7 & 38.6 & 36.2 & 38.2 & 52.8 \\
\midrule

\multicolumn{14}{c}{\textbf{Locally evaluated omni models}} \\
\midrule
ChronusOmni & AV & 13.1 & 37.3 & 13.8 & 24.2 & 22.7 & 26.0 & 41.6 & 26.2 & 25.0 & 15.7 & 23.4 & 33.4 \\
Qwen3-Omni & AV & 11.1 & 15.2 & 9.8 & 10.6 & 4.6 & 14.5 & 23.7 & 9.5 & 12.2 & 2.8 & 10.7 & 14.4 \\
Qwen2.5-Omni & AV & 5.1 & 7.1 & 5.8 & 5.1 & 6.1 & 7.3 & 8.1 & 2.9 & 5.8 & 1.4 & 5.1 & 7.3 \\
HumanOmniV2 & AV & 3.2 & 5.5 & 9.7 & 3.2 & 8.3 & 10.5 & 7.5 & 6.1 & 6.1 & 0.3 & 5.1 & 7.1 \\
Ola-7B & AV & 3.8 & 3.6 & 4.4 & 3.9 & 5.0 & 9.8 & 7.4 & 4.8 & 4.9 & 2.4 & 4.5 & 6.1 \\
MiniCPM-O-2.6 & AV & 1.3 & 1.0 & 2.3 & 1.4 & 2.5 & 3.7 & 4.2 & 1.5 & 2.0 & 3.3 & 2.2 & 2.7 \\
\midrule

\multicolumn{14}{c}{\textbf{Locally evaluated VL models}} \\
\midrule
TimeLens-8B & V & 19.7 & 44.1 & 26.5 & 29.3 & 27.0 & 43.9 & 43.7 & 29.3 & 31.9 & 21.3 & 30.1 & 41.7 \\
Qwen3-VL-8B & V & 14.5 & 41.0 & 24.3 & 21.8 & 26.8 & 47.6 & 38.4 & 25.0 & 28.3 & 16.4 & 26.3 & 35.0 \\
TimeLens-7B & V & 13.0 & 33.9 & 16.8 & 21.2 & 20.5 & 35.5 & 35.6 & 20.0 & 23.5 & 13.1 & 21.8 & 31.1 \\
Time-R1-7B & V & 14.4 & 28.5 & 16.7 & 19.0 & 15.0 & 25.5 & 34.3 & 20.2 & 21.1 & 12.7 & 19.7 & 29.4 \\
TRACE & V & 12.7 & 18.2 & 12.6 & 13.7 & 6.0 & 21.1 & 25.2 & 12.4 & 14.9 & 10.1 & 14.1 & 20.8 \\
Qwen2.5-VL-7B & V & 11.2 & 21.7 & 13.5 & 12.6 & 9.2 & 18.2 & 24.8 & 16.3 & 15.6 & 2.3 & 13.4 & 20.4 \\
VideoChat-R1 & V & 16.4 & 11.5 & 10.7 & 11.4 & 8.2 & 17.3 & 17.1 & 10.5 & 13.0 & 6.6 & 11.9 & 17.4 \\
TimeSuite & V & 5.5 & 13.8 & 6.6 & 8.2 & 10.9 & 15.1 & 20.5 & 9.4 & 10.6 & 3.6 & 9.4 & 13.8 \\
Video-R1-7B & V & 7.0 & 8.1 & 7.9 & 4.1 & 6.9 & 11.9 & 9.7 & 5.0 & 7.3 & 1.8 & 6.4 & 9.3 \\
MiMo-VL-7B-RL & V & 2.6 & 0.8 & 3.2 & 2.0 & 1.3 & 3.5 & 2.3 & 1.1 & 2.0 & 0.2 & 1.7 & 2.4 \\
\midrule

\multicolumn{14}{c}{\textbf{Human reference on the validated audit subset}} \\
\midrule
Human baseline & AV & 63.9 & 85.6 & 71.7 & 58.8 & 56.6 & 74.2 & 69.4 & 67.9 & 68.5 & 52.4 & 65.3 & 78.5 \\
\bottomrule
\end{tabular}%
}
\end{table*}

\subsection{Benchmark Analysis}
\label{sec:benchmark-analysis}

We first analyze three key properties that distinguish RVG-Bench from existing temporal grounding benchmarks.

\noindent\textbf{Existing benchmarks are approaching saturation.}
On the three TimeLens-optimized descriptive grounding benchmarks, namely Charades-TimeLens, ActivityNet-TimeLens, and QVHighlights-TimeLens, Gemini-2.5-Pro and TimeLens-8B---the strongest proprietary and open-source models evaluated in that study, respectively---achieve average mIoUs of 60.4 and 58.0~\cite{zhang2025timelens}. The gap between these leading models is only 2.5 points, or 4.2\% relative.
In contrast, Gemini-3-Pro and TimeLens-8B achieve 44.4 and 30.1 mIoU on RVG-Bench, respectively, with a substantially larger gap of 14.3 points, or 47.5\% relative to TimeLens-8B (\cref{tab:saturation_comparison}). This comparison motivates complementing descriptive grounding with reasoning-driven queries that remain challenging for strong models.

\noindent\textbf{RVG-Bench requires genuine audio-visual understanding.}
To examine modality reliance, we conduct modality ablation on three strong omni-modal models by removing audio to obtain V-only input or vision to obtain A-only input.
\cref{tab:modality_ablation_compact} summarizes the changes from A+V to V-only input across subsets with different modality requirements. Complete results, including A-only input and comparison with the traditional vision-centric Charades-TimeLens benchmark, are provided in \cref{tab:modality_ablation_comparison}.
Removing audio reduces Gemini-3-Pro's mIoU from 52.7 to 48.2 on the traditional benchmark and from 44.4 to 38.4 on RVG-Bench. On RVG-Bench, the drops are particularly pronounced for queries requiring audio or joint audio-visual evidence.
For example, under the V-only setting, Gemini-3-Pro drops from 53.1 to 24.9 on the Aud-Req subset and from 42.5 to 33.7 on the Both-Req subset.
These larger performance drops indicate that RVG-Bench contains many queries whose correct localization depends on modality-specific evidence beyond visual semantics, making it a more stringent test of omni-modal grounding.

\noindent\textbf{RVG-Bench is human-solvable.}
We conduct a stratified human audit on 200 examples sampled from the final human-validated benchmark. We sample 20 examples from each of the eight single-step reasoning types and 40 multi-step examples. Two annotators independently watch the complete video, read the query, and localize the temporal segment that directly answers it. For each example, the human baseline averages the two annotators' IoUs with the GT before aggregation. As shown in \cref{tab:main_miou}, humans achieve 68.5 mIoU on S-Step and 52.4 mIoU on M-Step; averaging over the 200 samples gives 65.3 mIoU and 78.5 IoU@0.3. Benefiting from our carefully designed construction pipeline and full human validation, RVG-Bench is a high-quality benchmark whose reasoning-driven grounding queries are reliably solvable by humans. Additional protocol details and aggregate results are provided in Appendix~\ref{sec:supp_human_baseline}.

\subsection{Model Analysis}
\label{sec:model-analysis}

Having established the benchmark properties, we next analyze the capabilities of current models. 
Overall results in \cref{tab:main_miou} indicate that RVG-Bench poses substantial challenges for existing systems. 
Qwen3.8-Omni-Flash leads with 49.6 mIoU and 68.0 IoU@0.3, followed by Seed2.0-Lite at 47.0 mIoU and Gemini-3-Pro at 44.4. 
The leading model remains below the human reference of 65.3 mIoU and 78.5 IoU@0.3 on the 200-example audit subset.
Among locally evaluated models, TimeLens-8B leads with 30.1 mIoU and 41.7 IoU@0.3. 
The strongest local baselines are trained for temporal localization. The evaluated local omni models still lag behind these grounding models despite receiving both audio and video, suggesting that access to both modalities alone does not ensure strong reasoning-driven grounding.

Aggregate metrics alone do not identify whether failures arise from perception, target inference, or temporal alignment. We examine these factors through oracle descriptions and prompting strategies below. Complementary diagnoses by two judges rank localization as the most frequent error category for Gemini-3-Pro and Qwen3-VL-8B when pooled across the three settings (Appendix~\ref{sec:judge_agreement}).

\begin{table}[t]
\centering
\footnotesize
\setlength{\tabcolsep}{3.2pt}
\caption{\textbf{Compact modality ablation on RVG-Bench.} Entries show A+V $\rightarrow$ V-only mIoU over all single- and multi-step queries in the evaluation subset.}
\label{tab:modality_ablation_compact}
\begin{tabular}{lccc}
\toprule
\textbf{Model} & \textbf{Vis-Req} & \textbf{Aud-Req} & \textbf{Both-Req} \\
\midrule
ChronusOmni  & 25.6$\rightarrow$23.8 & 23.7$\rightarrow$18.2 & 21.1$\rightarrow$18.3 \\
Gemini-3-Pro & 44.0$\rightarrow$46.2 & 53.1$\rightarrow$24.9 & 42.5$\rightarrow$33.7 \\
Qwen3-Omni   & 8.8$\rightarrow$10.9 & 15.9$\rightarrow$10.0 & 11.2$\rightarrow$10.2 \\
\bottomrule
\end{tabular}
\end{table}

\begin{figure*}[t!]
  \centering
  \begin{subfigure}[b]{0.48\textwidth}
    \centering
    \resizebox{\linewidth}{!}{%
      \begin{tabular}{lcc}
        \toprule
        \textbf{Model} & \textbf{Default} & \textbf{Ground} \\
        \midrule
        Qwen3.8-Omni-Flash & 49.6 & 59.1$_{+9.5}$ \\
        Seed2.0-Lite & 47.0 & 57.6$_{+10.6}$ \\
        Gemini-3-Pro & 44.4 & 63.9$_{+19.5}$ \\
        ChronusOmni & 23.4 & 35.2$_{+11.7}$ \\
        TimeLens-8B & 30.1 & 45.4$_{+15.3}$ \\
        \midrule
        \textbf{Avg. 15 models} & 26.9 & 36.8$_{+9.9}$ \\
        \bottomrule
      \end{tabular}%
    }
    \caption{Oracle grounding for Overall mIoU.}
    \label{fig:oracle_grounding_trend}
  \end{subfigure}
  \hfill
  \begin{subfigure}[b]{0.48\textwidth}
    \centering
    \includegraphics[width=\linewidth]{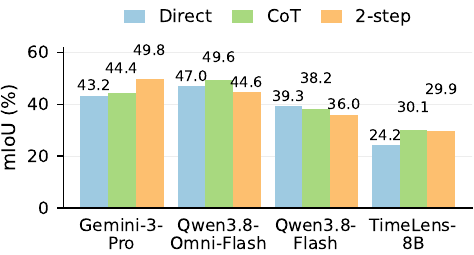}
    \caption{Task decomposition for Overall mIoU.}
    \label{fig:task_decomposition_trend}
  \end{subfigure}
  \caption{Compact results for representative model-analysis settings. In (a), subscripts denote Ground $-$ Default. Complete Oracle-grounding and task-decomposition results are reported in \cref{tab:full_default_vs_ground_miou,tab:direct_cot_2step_miou_summary}.}
  \label{fig:compact_model_analysis}
\end{figure*}

\noindent\textbf{Fine-grained analysis across reasoning categories.}
We report performance across eight single-step types and the multi-step category in \cref{tab:main_miou}. Performance varies substantially across reasoning types. For most models, causal queries are among the most difficult single-step types: Gemini-3-Pro reaches 31.3 mIoU, while TimeLens-8B reaches 19.7.
Conversely, models generally perform best on \textit{Seq} and \textit{Comp} categories. This gap suggests that identifying implicit causal relations involving subtle and temporally separated cues remains a major challenge. 
Multi-step queries further amplify this difficulty. 
For instance, Gemini-3-Pro drops from 46.1 mIoU on single-step queries to 35.8 on multi-step queries, and TimeLens-8B drops from 31.9 to 21.3. Qwen3-Omni and Qwen2.5-VL-7B score only 2.8 and 2.3 on multi-step queries, respectively. These results highlight the difficulty of maintaining coherent long-horizon reasoning in untrimmed videos.


\noindent\textbf{Disentangling reasoning from localization.} We compare end-to-end grounding with localization given an oracle target description to examine reasoning about \emph{what} to locate and determining \emph{when} it occurs. Each query is paired with its ground-truth textual answer describing the target content, and the model localizes the corresponding span.
In \cref{fig:oracle_grounding_trend}, all 15 models in the analysis set improve in overall mIoU, with the mean rising from 26.9 to 36.8. Qwen3.8-Omni-Flash and Seed2.0-Lite rise from 49.6 and 47.0 to 59.1 and 57.6, while Gemini-3-Pro reaches 63.9, gaining 19.5 points. ChronusOmni and TimeLens-8B reach 35.2 and 45.4. The 22-model comparison in \cref{tab:full_default_vs_ground_miou} shows an average gain of 8.4 points. These overall gains reflect the benefit of making the target explicit, including the richer semantic retrieval cues provided by its description. However, grounding remains difficult for models with limited localization ability: even with the correct target supplied, Qwen2.5-Omni and MiniCPM-O-2.6 reach only 9.0 and 4.3 mIoU. These results motivate improving temporal localization alongside audio-visual reasoning about the target content.

\noindent\textbf{Scaling via task decomposition.} The gains from oracle target descriptions raise a practical question: can more explicit reasoning or task decomposition improve end-to-end grounding without access to ground-truth answers? We compare \textbf{Direct}, which predicts timestamps directly; \textbf{CoT}, which generates reasoning and timestamps in one pass; and \textbf{2-step}, which first answers the query and then localizes the generated target description.
Among these strategies, none consistently dominates across models and query types. In \cref{fig:task_decomposition_trend}, 2-step performs best for Gemini-3-Pro (49.8 mIoU), CoT for Qwen3.8-Omni-Flash (49.6), and Direct for Qwen3.8-Flash (39.3), while TimeLens-8B performs similarly under CoT and 2-step (30.1 vs.\ 29.9). The preferred strategy can also change within a model: Qwen3.8-Flash achieves its highest single-step score with Direct (40.1), but its highest multi-step score with 2-step (37.6; Appendix~\ref{sec:supp_full_tts}). Thus, adding explicit reasoning or separating answer generation from localization does not guarantee better grounding. Together with the remaining errors under oracle grounding, these findings motivate strengthening models' ability to infer targets from audiovisual evidence and localize them precisely, alongside inference strategies adapted to different models and queries.

\vspace{-2mm}
\section{Conclusion}
\label{sec:conclusion}
We introduced RVG-Bench, a benchmark that shifts temporal video grounding from surface-level descriptive matching toward reasoning-driven localization. By requiring models to infer implicit targets through diverse reasoning types over joint audio-visual evidence, RVG-Bench exposes limitations of current models that existing benchmarks largely overlook.  
Through systematic analyses, including fine-grained category breakdowns, oracle-grounding probes, and test-time scaling, we identify challenges in both target inference and temporal localization. 
We hope RVG-Bench will serve as a challenging testbed for advancing multimodal models with stronger audio-visual perception and temporal reasoning capabilities.

\vspace{-2mm}
\section*{Limitations}
\label{sec:limitations}

We acknowledge that RVG-Bench currently focuses on short-to-medium length videos, leaving extensively long video sources such as full-length movies or hour-long uncut footage unexplored. Consequently, the benchmark does not evaluate movie-level narrative understanding or long-term memory capabilities. However, because reasoning-driven temporal grounding is a newly introduced and highly demanding task formulation, our empirical results demonstrate that current state-of-the-art multimodal models still struggle significantly even on these shorter videos. Establishing a rigorous baseline on short-to-medium videos thus serves as a necessary first step for the community. Extending reasoning-driven video grounding to long-form content is an important frontier that we leave for future exploration.

\section*{Acknowledgments and Disclosure of Funding}
This work was supported in part by the Beijing Major Science and Technology Project under Contract no.~Z251100008425023.

\bibliographystyle{splncs04}
\bibliography{main}

\clearpage
\appendix
\renewcommand{\thesubsection}{\thesection.\arabic{subsection}}
\renewcommand{\thesubsubsection}{\thesubsection.\arabic{subsubsection}}
\renewcommand{\theHsection}{supp.\arabic{section}}
\renewcommand{\theHfigure}{supp.\arabic{figure}}
\renewcommand{\theHtable}{supp.\arabic{table}}
\renewcommand{\theHequation}{supp.\arabic{equation}}

\section*{Appendix}
\suppressfloats[t]

\section{Benchmark Details and Quality Validation}
\label{sec:supp_benchmark_details}

This section details the benchmark design, statistics, human validation, human baseline, and qualitative cases. Appendix~\ref{sec:supp_eval_details} documents the evaluation protocol, Appendix~\ref{sec:extended_main_analyses} reports experimental results and analyses, and Appendix~\ref{sec:supp_prompt_templates} presents the prompt templates.

\subsection{Comparison with Representative Benchmarks}
\label{sec:supp_benchmark_comparison}

\newcommand{\cmark}{\ding{51}} 
\newcommand{\xmark}{\ding{55}} 

\begin{table*}[ht!]
\centering
\caption{Comparison with representative TVG benchmarks. For datasets that provide both training and test splits, we report statistics on the \textbf{test split only}. \textbf{Audio} indicates whether queries require audio information. \textbf{Rel} indicates relation reasoning. \textbf{Attr} indicates attribute reasoning, including both intrinsic attributes and relative attributes. \textbf{Multi-step} indicates whether the benchmark contains multi-step reasoning queries. The symbol ``--'' indicates that the specific statistics are neither reported in their original paper nor available because the dataset is not open source.}
\label{tab:benchmark_comparison}
\small
\setlength{\tabcolsep}{4pt}
\resizebox{\textwidth}{!}{%
\begin{tabular}{lccccccc}
\toprule
\textbf{Benchmark} &
\textbf{\#Videos} &
\textbf{Avg.\ len (s)} &
\textbf{\#Queries} &
\textbf{Audio} &
\textbf{Rel} &
\textbf{Attr} &
\textbf{Multi-step} \\
\midrule
Charades-STA~\cite{sigurdsson2016hollywood}    & 1,334 & 29.5  & 3,720  & \xmark & \xmark & \xmark & \xmark \\
ActivityNet Caps~\cite{krishna2017dense} & 4,885 & 118.1 & 17K    & \xmark & \xmark & \xmark & \xmark \\
QVHighlights~\cite{lei2021detecting}     & 1,529 & 149.6 & 1,542  & \xmark & \xmark & \xmark & \xmark \\
LongVALE~\cite{geng2025longvale}         & 1,171 & 235.0 & 14K    & \cmark & \xmark & \xmark & \xmark \\
ChronusAV~\cite{chen2025chronusomni}        & 2,000 & 226.0 & 12K    & \cmark & \xmark & \xmark & \xmark \\
ToTG-Bench~\cite{liu2025timescope}  & -- & -- & -- & \xmark & \cmark & \xmark & \xmark \\
UnAV-100~\cite{geng2023dense}        & 2,167 & 42.1  & 5,900  & \cmark & \xmark & \xmark & \xmark \\
Shot2Story~\cite{han2025shot2story}  & 4,025 & 17.1  & 6,465  & \xmark & \xmark & \xmark & \xmark \\
AVEL~\cite{tian2018audio}            & 402   & 10.0  & 402    & \cmark & \xmark & \xmark & \xmark \\
Ego4D-NLQ~\cite{grauman2024ego4d}    & 300   & 552.0 & 4,000  & \xmark & \xmark & \xmark & \xmark \\
\midrule
\textbf{RVG-Bench} &
\textbf{1,255} & \textbf{108.4} & \textbf{1,890} &
\textbf{\cmark} & \textbf{\cmark} & \textbf{\cmark} & \textbf{\cmark} \\
\bottomrule
\end{tabular}%
}

\end{table*}

\FloatBarrier

\subsection{Full Reasoning Taxonomy with Examples}
\label{sec:supp_full_taxonomy}

\begin{table*}[t!]
\centering
\small
\caption{\textbf{Full reasoning taxonomy of RVG-Bench.} Each subcategory is presented in two rows: one defining how the query indirectly specifies the target and one providing an example query. Subcategory abbreviations follow \cref{tab:taxonomy_compact}.}
\label{tab:taxonomy_full}
\setlength{\tabcolsep}{3pt}
\renewcommand{\arraystretch}{1.08}
\begin{tabular}{P{1.8cm} M{1.4cm} M{1.45cm} P{6.5cm}}
\toprule
\textbf{\makecell{Major\\Category}} & \textbf{\makecell{Sub-\\category}} & \textbf{Entry} & \textbf{Content} \\
\midrule
\multirow{6}{=}{\makecell{Intrinsic\\Attribute}}
& \multirow{2}{*}{Char} & Definition & Identify a segment through a character's implicit mental state or behavioral pattern. \\
& & Example & \emph{``Which segment shows the girl's hesitation?''} \\ \cmidrule{2-4}
& \multirow{2}{*}{Anom} & Definition & Locate a segment that deviates from the expected norm of surrounding context. \\
& & Example & \emph{``Locate the abnormal moment.''} \\ \cmidrule{2-4}
& \multirow{2}{*}{Symb} & Definition & Interpret abstract or metaphorical meaning to identify the target. \\
& & Example & \emph{``Locate the scene that symbolizes freedom.''} \\
\midrule
\multirow{4}{=}{\makecell{Relative\\Attribute}}
& \multirow{2}{*}{Comp} & Definition & Select a segment by comparing a measurable property across candidates. \\
& & Example & \emph{``Which jump is the highest?''} \\ \cmidrule{2-4}
& \multirow{2}{*}{Ord} & Definition & Locate a segment by its position in a sequence. \\
& & Example & \emph{``Locate the last appearance of the boy doing high jump.''} \\
\midrule
\multirow{6}{=}{Relational}
& \multirow{2}{*}{Caus} & Definition & Trace from an effect to its cause, or vice versa. \\
& & Example & \emph{``Which segment causes the glass to break?''} \\ \cmidrule{2-4}
& \multirow{2}{*}{Seq} & Definition & Identify a segment by temporal adjacency to a specified event. \\
& & Example & \emph{``What happens right before adding salt?''} \\ \cmidrule{2-4}
& \multirow{2}{*}{Ref} & Definition & Resolve an anaphoric reference to locate the referred segment. \\
& & Example & \emph{``Locate the segment where `he' in `He is here' appears.''} \\
\midrule
\multirow{2}{=}{Multi-step}
& \multirow{2}{*}{M-Step} & Definition & Compose two or more atomic reasoning skills within a single query. \\
& & Example & \emph{``Which segment reveals why she hesitates?''} \\
\bottomrule
\end{tabular}
\end{table*}

\FloatBarrier

\subsection{Benchmark Statistics and Distributions}
\label{sec:supp_stats}

\begin{figure*}[t!]
  \centering
  \captionsetup[subfigure]{%
    font=footnotesize,
    justification=centering,
    singlelinecheck=true
  }
  \hyphenation{distribution position normalized}

  \begin{subfigure}[t]{0.32\textwidth}
    \centering
    \includegraphics[width=\linewidth]{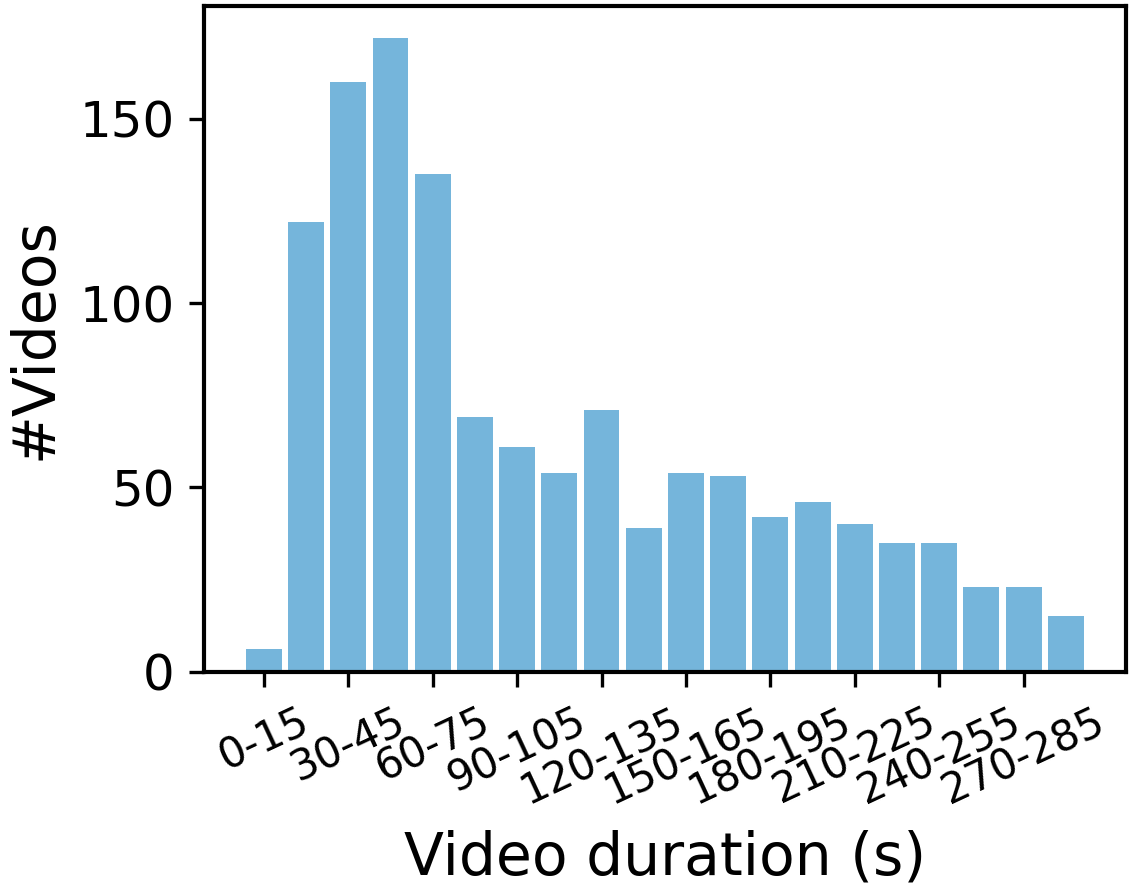}
    \caption{Video duration distribution.}
    \label{fig:stats-vlen}
  \end{subfigure}\hfill
  \begin{subfigure}[t]{0.32\textwidth}
    \centering
    \includegraphics[width=\linewidth]{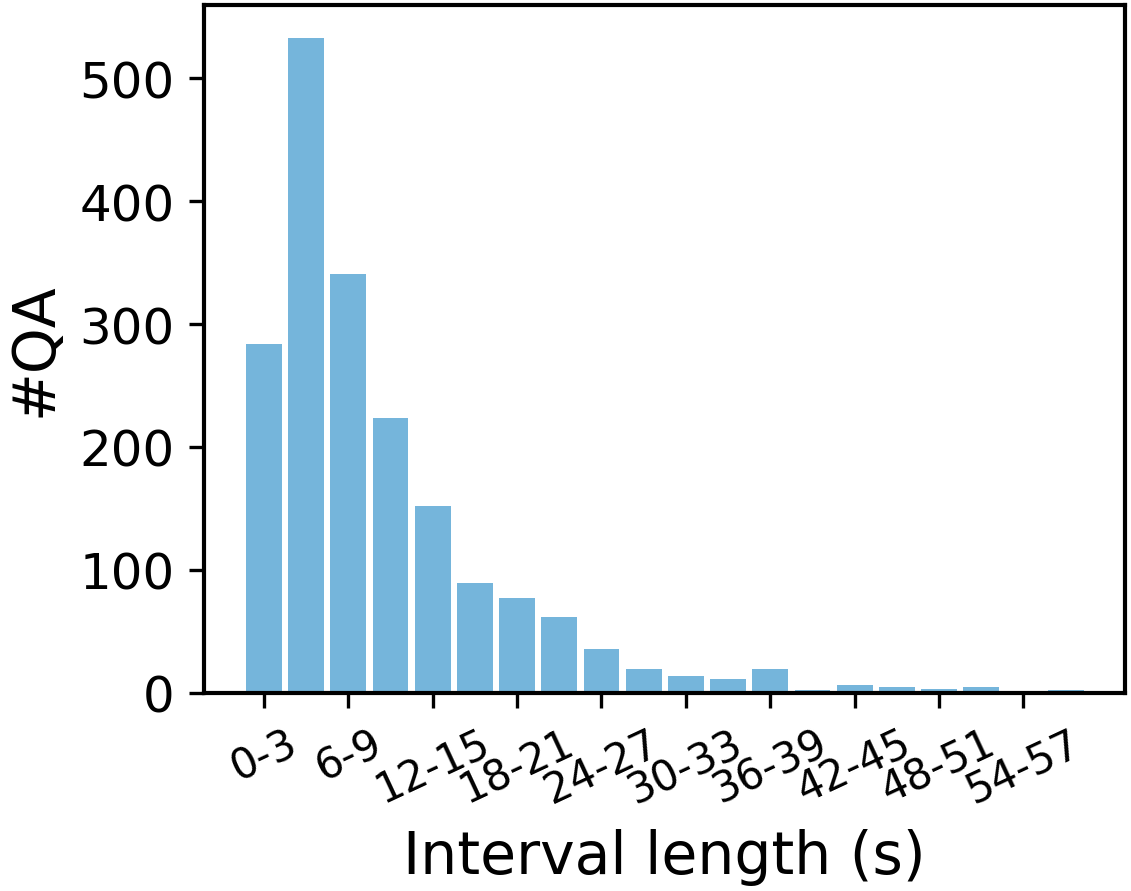}
    \caption{GT interval length distribution.}
    \label{fig:stats-gtlen}
  \end{subfigure}\hfill
  \begin{subfigure}[t]{0.32\textwidth}
    \centering
    \includegraphics[width=\linewidth]{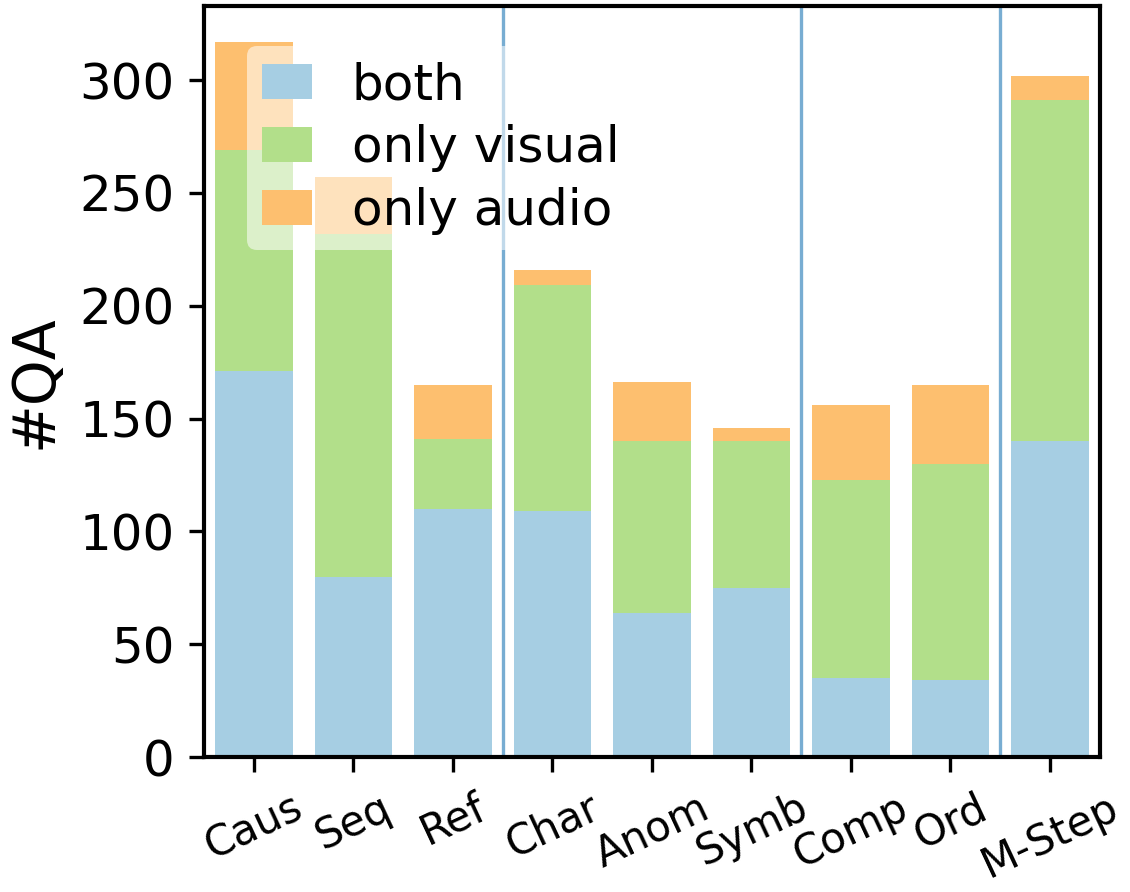}
    \caption{Reasoning category and modality.}
    \label{fig:stats-catmod}
  \end{subfigure}

  \vspace{2pt}
  \begin{subfigure}[t]{0.32\textwidth}
    \centering
    \includegraphics[width=\linewidth]{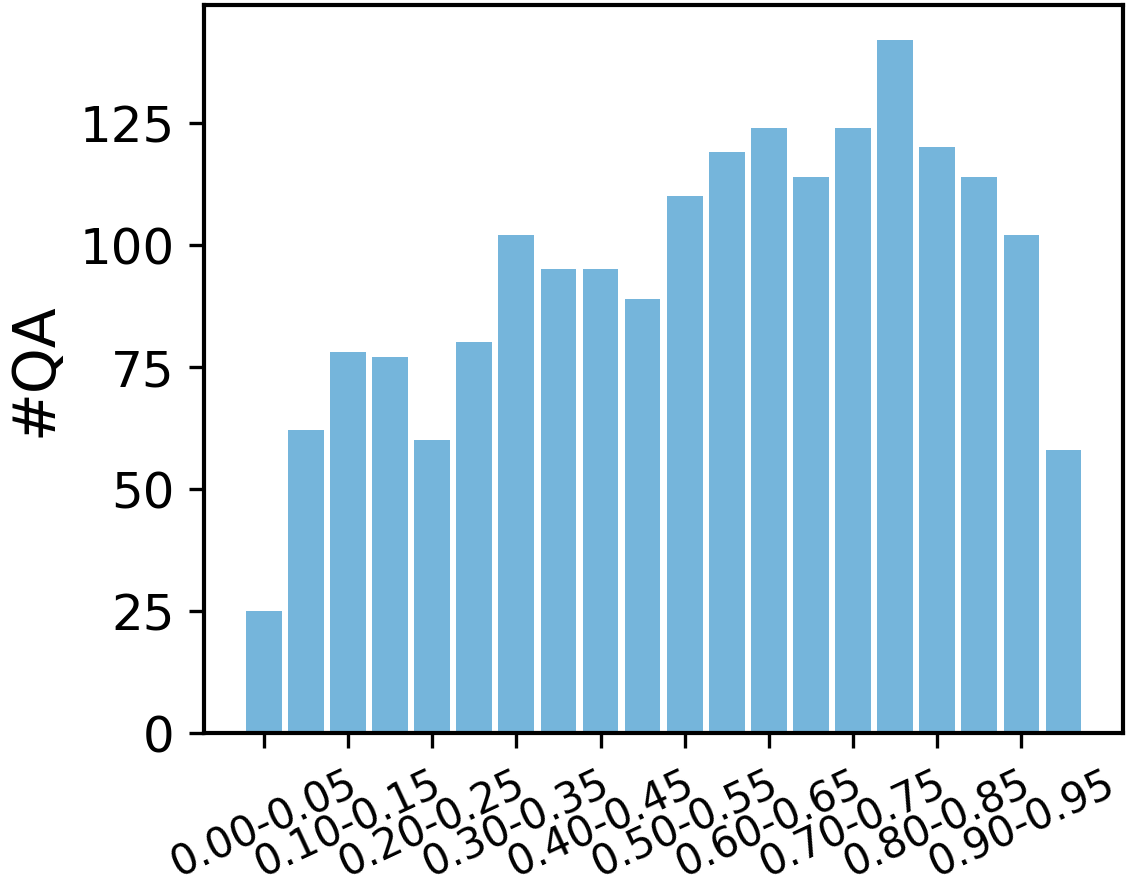}
    \caption{Normalized center position.}
    \label{fig:gt-center}
  \end{subfigure}\hfill
  \begin{subfigure}[t]{0.32\textwidth}
    \centering
    \includegraphics[width=\linewidth]{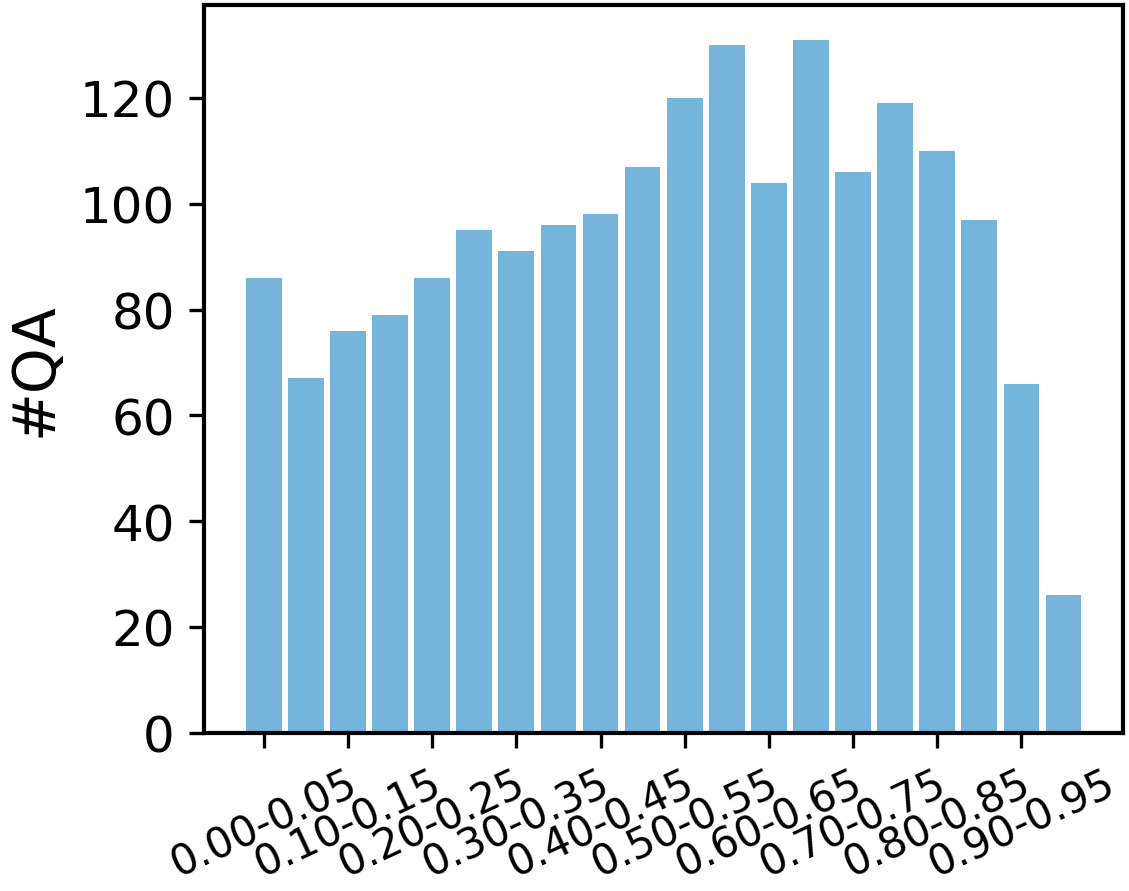}
    \caption{Normalized start position.}
    \label{fig:gt-start}
  \end{subfigure}\hfill
  \begin{subfigure}[t]{0.32\textwidth}
    \centering
    \includegraphics[width=\linewidth]{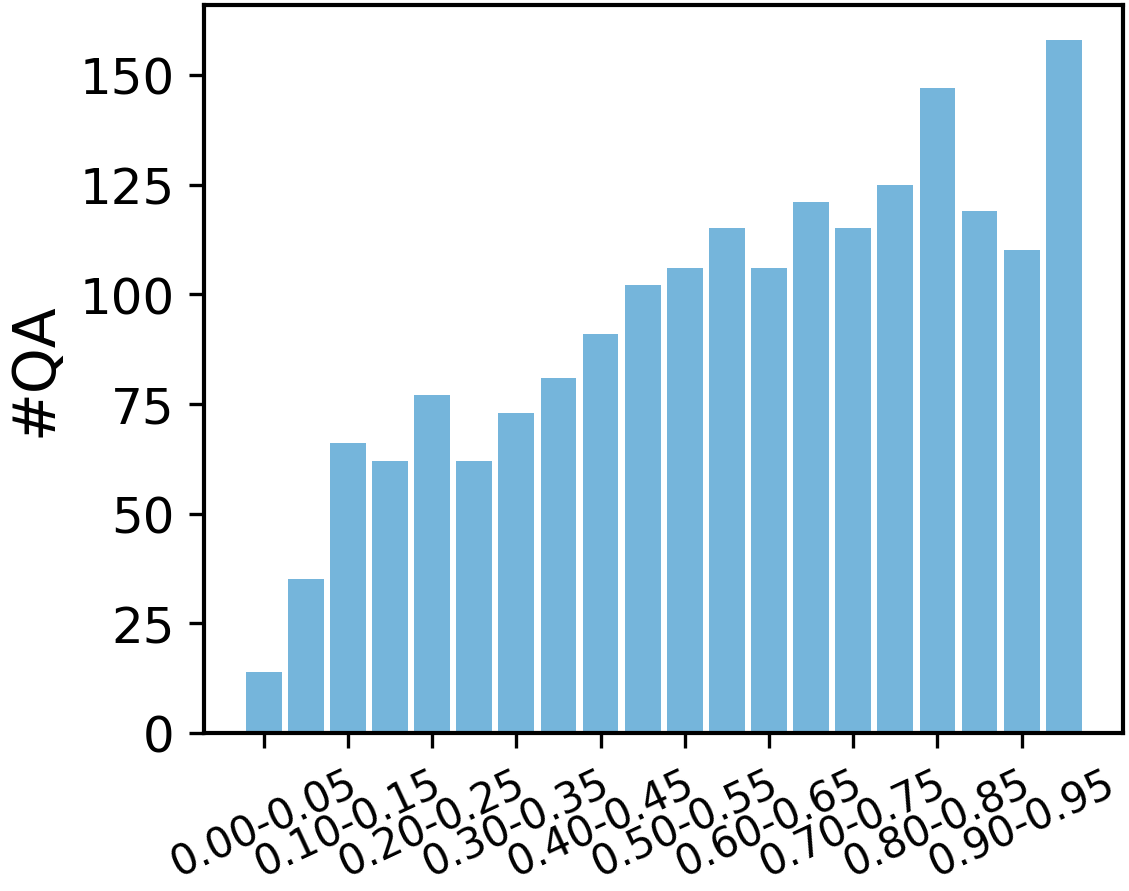}
    \caption{Normalized end position.}
    \label{fig:gt-end}
  \end{subfigure}
  \vspace{-8pt}
  \caption{Statistics of RVG-Bench, including video duration, ground-truth (GT) interval length, reasoning-category distribution with required modality, and normalized temporal distributions of the center, start, and end points of GT intervals.}
  \label{fig:benchmark_statistics}
  \label{fig:stats-length}
  \label{fig:gt_position_dist}
\end{figure*}

As discussed in the main manuscript, it is crucial to ensure that the dataset does not suffer from temporal positional biases. If the target segments were heavily concentrated at specific parts of the videos, such as the middle or the end, models could exploit these statistical shortcuts to achieve artificially high scores without genuinely performing multi-step reasoning.

To examine temporal positional biases, Figure~\ref{fig:gt_position_dist} illustrates the normalized temporal distribution of the ground-truth segments. We separately plot the distributions of the center, start, and end points of the target intervals relative to the total video duration. The broad coverage of the timeline reduces the usefulness of simple temporal-position priors.

\FloatBarrier

\subsection{Full Human Validation}
\label{sec:supp_human_validation}

To ensure that the automated pipeline produces reliable and high-quality annotations, all 2,500 candidates produced by Stage~3 underwent full human review using a dedicated web-based annotation interface. Each candidate was reviewed by one graduate-student annotator. This validation is distinct from the human-baseline study in \cref{sec:supp_human_baseline}, which uses two independent annotations per example. Across this process, 610 samples, accounting for 24.40\%, were removed and 1,890 were retained.

\begin{figure*}[t!]
  \centering
  \includegraphics[width=0.9\linewidth]{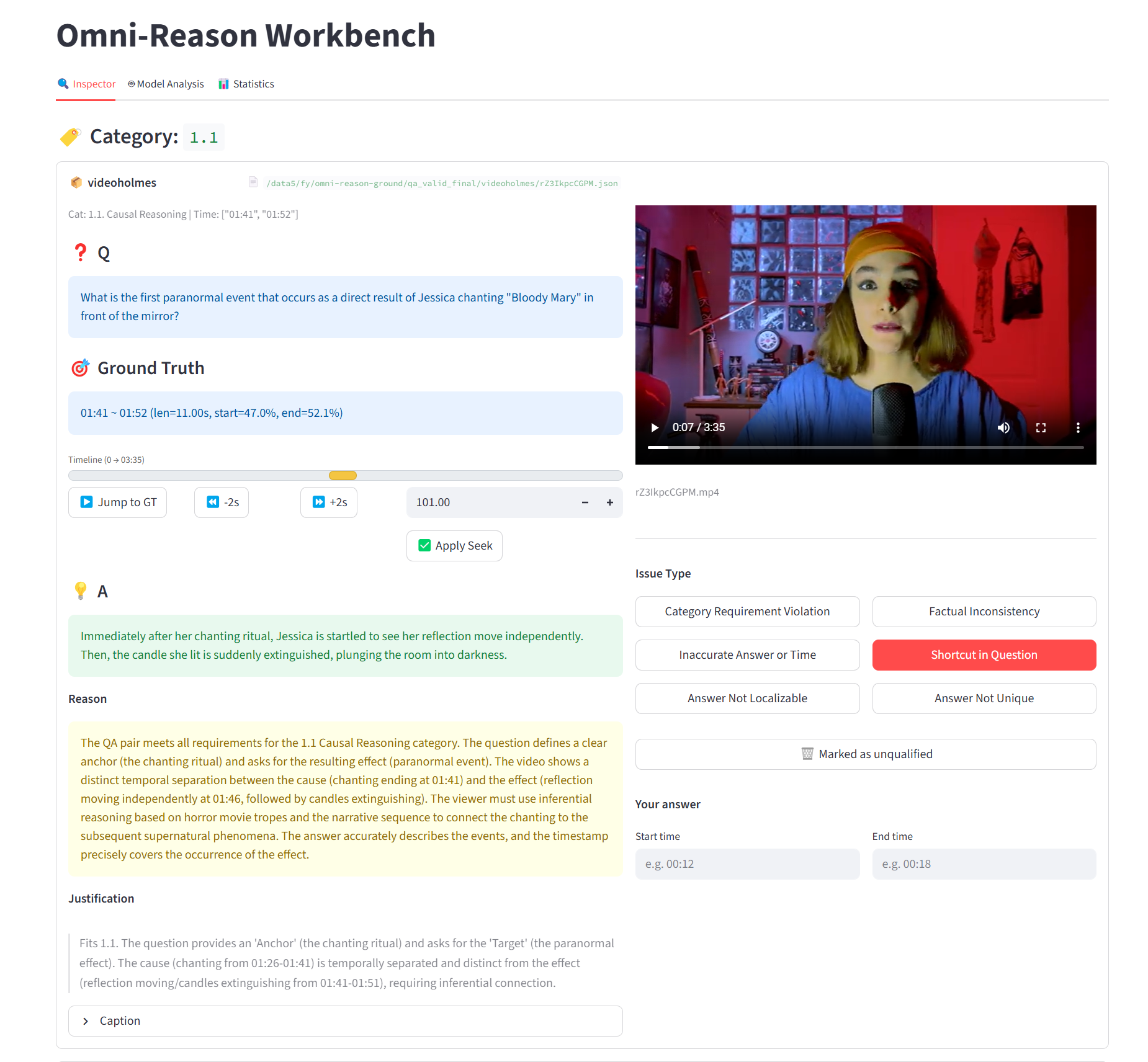}
  \vspace{-2mm}
  \caption{Screenshot of our human validation interface. Annotators can flag six issue types: \textit{Category Requirement Violation}, \textit{Factual Inconsistency}, \textit{Inaccurate Answer or Time}, \textit{Shortcut in Question}, \textit{Answer Not Localizable}, and \textit{Answer Not Unique}. For reporting, category violations and factual inconsistencies are assessed under \textbf{Logical soundness}; answer correctness and localizability under \textbf{Answerability}; shortcuts under \textbf{No information leakage}; and non-unique targets under \textbf{Uniqueness}. Temporal-boundary refinements are recorded separately. Issue types may overlap.}
  \label{fig:human_validation_ui}
\end{figure*}

As illustrated in \cref{fig:human_validation_ui}, the validation protocol strictly requires human annotators to execute the following steps:

\begin{enumerate}
    \item \textbf{Context Familiarization:} Annotators are required to watch the full video from start to finish to comprehensively grasp the storyline and contextual details.
    \item \textbf{Query and Answer Quality Assessment:} Based on the video content, annotators evaluate the generated query-answer pair against four strict criteria:
    \begin{itemize}
        \item \textit{Answerability:} Is the query answerable from the video, and is the provided answer correct and responsive to the query?
        \item \textit{Uniqueness:} Is there only one definitive segment in the video that satisfies the query?
        \item \textit{Logical soundness:} Is the reasoning path valid, aligned with the designated category, and factually consistent with the audio-visual evidence?
        \item \textit{No information leakage:} Is the query free of surface-level shortcuts that directly reveal the target segment?
    \end{itemize}
    \item \textbf{Question and Temporal-Grounding Refinement:} Samples with unresolved quality issues are removed. For otherwise valid samples, annotators rewrite questions with wording problems and refine the GT start and end timestamps whenever the original interval does not precisely cover the target event.
\end{enumerate}

\Cref{tab:human_validation_pass_rates} reports the initial pass rate for each criterion over all 2,500 candidates, based on the original questions and answers. Each criterion is assessed independently: a query--answer pair may fail multiple criteria simultaneously, so the criterion-specific failure counts are not mutually exclusive. Final selection and refinement outcomes are shown in a separate block of \cref{tab:human_validation_stats}.

\begin{table*}[t!]
\centering
\caption{Initial criterion-wise pass rates and final validation outcomes. Counts in the first block denote passed samples. The denominator for each block is shown explicitly.}
\label{tab:human_validation_stats}
\label{tab:human_validation_pass_rates}
\small
\setlength{\tabcolsep}{8pt}
\begin{tabular}{@{}lrr@{}}
\toprule
\textbf{Criterion / outcome} & \textbf{Count} & \textbf{Rate} \\
\midrule
\multicolumn{3}{@{}l}{\textit{Initial criterion-wise assessment ($n=2{,}500$)}} \\
Answerability & 2,091 & 83.64\% \\
Uniqueness & 2,447 & 97.88\% \\
Logical soundness & 2,045 & 81.80\% \\
No information leakage & 2,480 & 99.20\% \\
\midrule
\multicolumn{3}{@{}l}{\textit{Final selection ($n=2{,}500$)}} \\
Retained & 1,890 & 75.60\% \\
Removed & 610 & 24.40\% \\
\midrule
\multicolumn{3}{@{}l}{\textit{Refinement among retained samples ($n=1{,}890$)}} \\
GT timestamps refined & 240 & 12.70\% \\
Questions rewritten & 98 & 5.19\% \\
\bottomrule
\end{tabular}
\end{table*}

Among the 1,890 retained samples, annotators refined the GT timestamps for 240 samples, accounting for 12.70\%, and questions in 98 samples, accounting for 5.19\%, were rewritten. These operations produce the final benchmark used throughout the experiments.

\FloatBarrier

\subsection{Human Baseline}
\label{sec:supp_human_baseline}

We additionally measure human temporal-grounding performance on a stratified subset of 200 examples sampled from the final human-validated benchmark. We sample 20 examples from each of the eight single-step reasoning types and 40 examples from the multi-step category. Every example is independently annotated by two annotators. Each annotator watches the full video, reads the query, and marks the start and end timestamps of the segment that directly answers it. The human baseline averages the two annotators' IoUs with the GT for each query before aggregating over all 200 samples. As shown in \cref{tab:human_baseline}, humans achieve 65.3 mIoU and 78.5 IoU@0.3. Human--human agreement reaches 73.7 mIoU, with IoU@0.3, IoU@0.5, and IoU@0.7 scores of 80.0, 77.5, and 60.0, respectively. These results support the temporal stability and human solvability of the final benchmark.

\begin{table}[t!]
\centering
\small
\setlength{\tabcolsep}{4pt}
\caption{Human temporal-grounding results on the 200-example double-annotated audit subset. The human baseline averages the two annotators' IoU with GT per query. Human--human agreement compares the two annotators directly. All metric values are percentages.}
\label{tab:human_baseline}
\resizebox{\columnwidth}{!}{%
\begin{tabular}{lrrrrr}
\toprule
\textbf{Reference} & $N$ & \textbf{mIoU} & \textbf{@0.3} & \textbf{@0.5} & \textbf{@0.7} \\
\midrule
Human baseline (vs. GT) & 200 & 65.3 & 78.5 & 74.0 & 55.5 \\
Human--human agreement & 200 & 73.7 & 80.0 & 77.5 & 60.0 \\
\bottomrule
\end{tabular}%
}
\end{table}

\FloatBarrier

\subsection{Additional Benchmark Cases}
\label{sec:supp_more_cases}

To provide a more intuitive understanding of the reasoning taxonomy defined in the main manuscript, we present comprehensive qualitative examples for each major reasoning category. These examples illustrate how RVG-Bench moves beyond conventional surface-level descriptive grounding to require deep, indirect inference across various video contexts.

\begin{figure*}[htbp]
  \centering
  \includegraphics[width=0.95\linewidth]{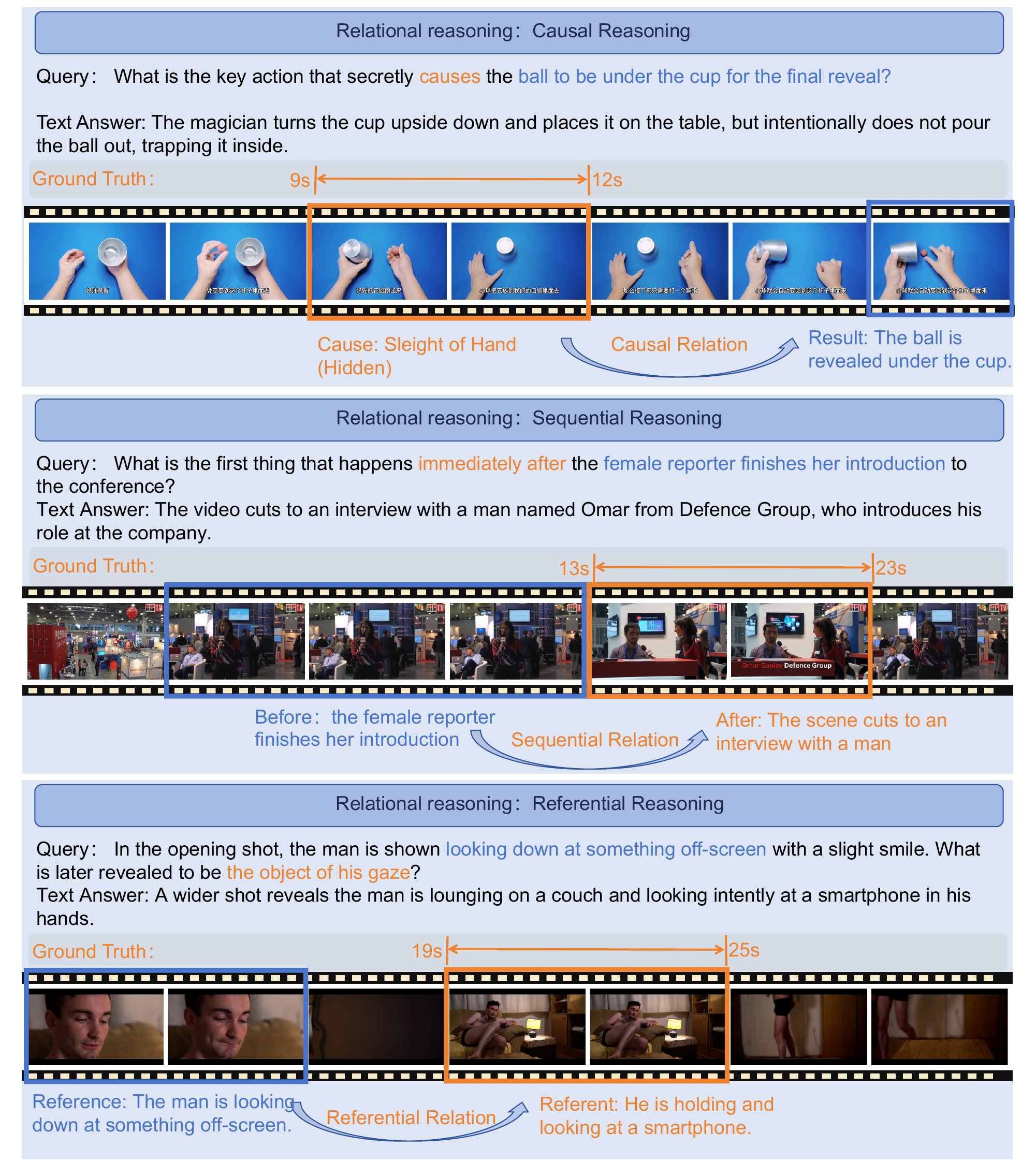}
  \vspace{-5pt}
  \caption{\textbf{Examples of Relational reasoning.} The figure encompasses three cases covering the \textit{Causal}, \textit{Sequential}, and \textit{Referential} subcategories. Success in these tasks hinges on the model's ability to navigate structured dependencies within the video.}
  \label{fig:supp_case_relational}
\end{figure*}

\noindent\textbf{Relational Reasoning.} 
Figure~\ref{fig:supp_case_relational} presents examples from the relational reasoning category. Videos inherently possess structured narrative dependencies. These queries specify one element of a structured relationship and demand the model to traverse causal chains, temporal sequences, or visual/audio references to locate the corresponding counterpart.

\begin{figure*}[htbp]
  \centering
  \includegraphics[width=0.95\linewidth]{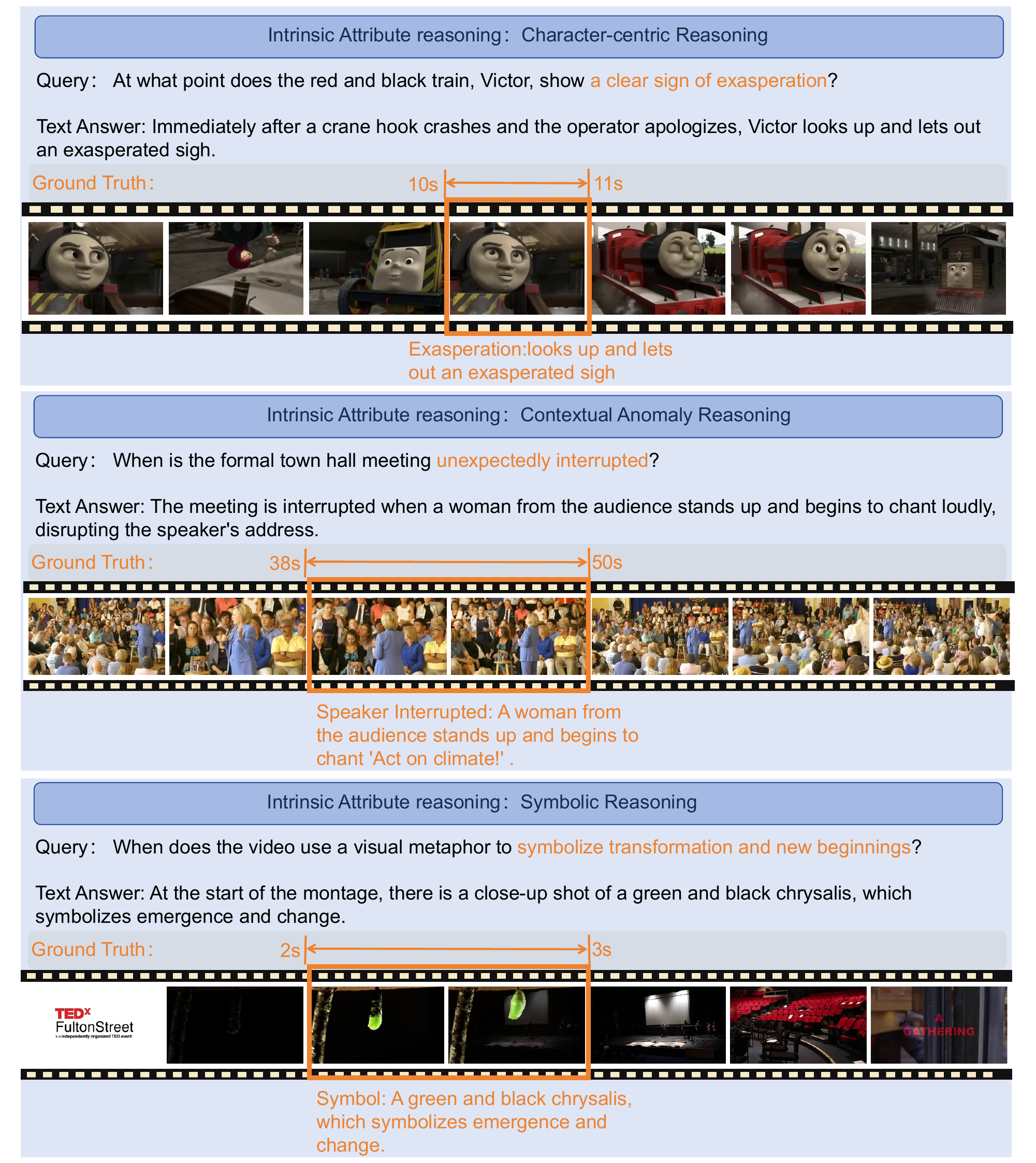}
  \vspace{-5pt}
  \caption{\textbf{Examples of Intrinsic-attribute reasoning.} The figure includes three distinct cases corresponding to the \textit{Character-centric}, \textit{Contextual Anomaly}, and \textit{Symbolic} subcategories. Models must reason over the implicit meaning or context of the video segments rather than observable surface features.}
  \label{fig:supp_case_intrinsic}
\end{figure*}

\noindent\textbf{Intrinsic-Attribute Reasoning.} 
Figure~\ref{fig:supp_case_intrinsic} displays three examples from the intrinsic-attribute category. In these cases, the queries refer to implicit properties such as a character's mental state, an anomalous occurrence, or a symbolic representation. The model cannot rely on direct visual or audio matching; instead, it must infer the underlying semantics of the scene to accurately localize the target segments.

\begin{figure*}[htbp]
  \centering
  \includegraphics[width=0.95\linewidth]{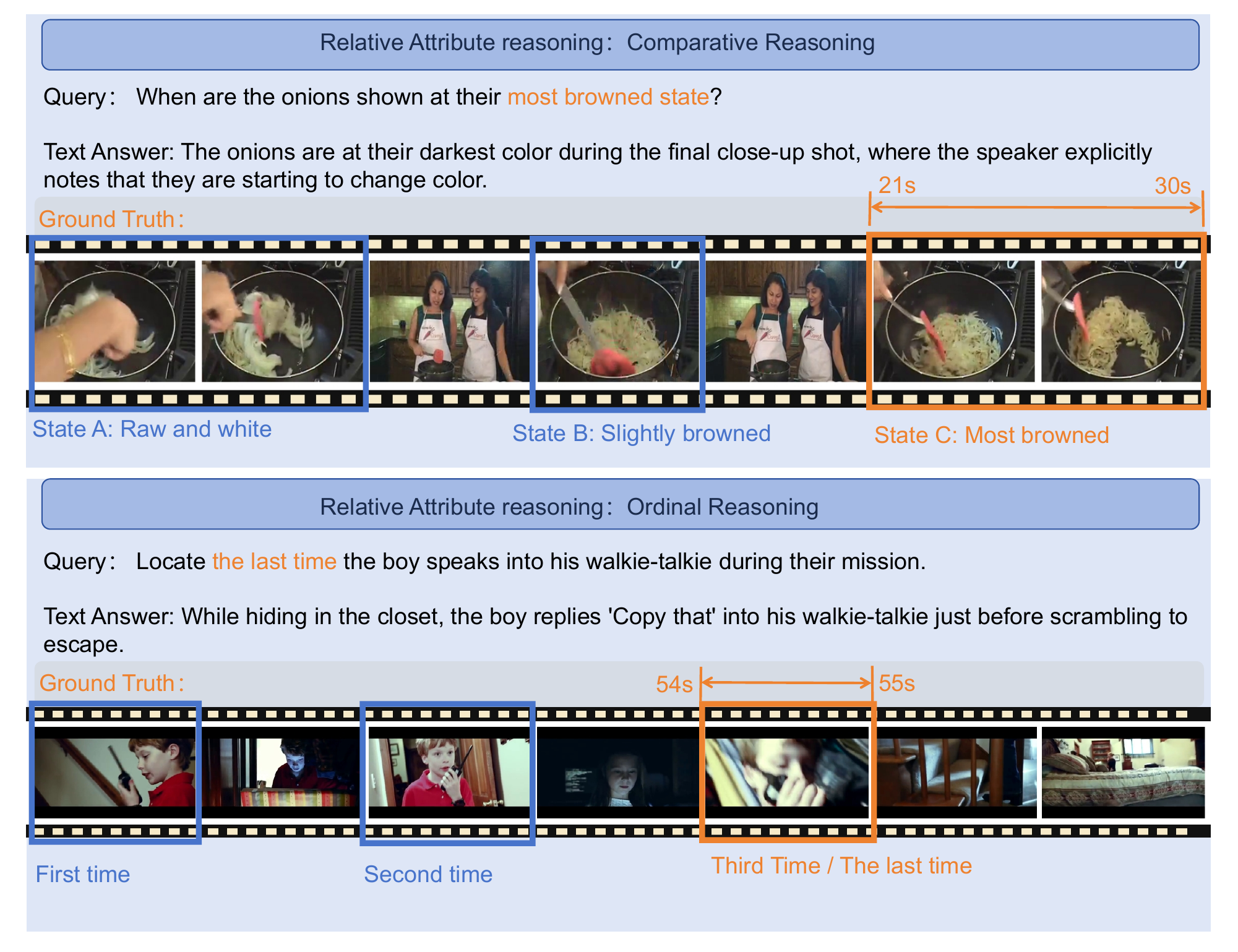}
  \vspace{-5pt}
  \caption{\textbf{Examples of Relative-attribute reasoning.} The two cases represent the \textit{Comparative} and \textit{Ordinal} subcategories, requiring the model to evaluate and rank multiple candidate segments across the entire video timeline.}
  \label{fig:supp_case_relative}
\end{figure*}

\noindent\textbf{Relative-Attribute Reasoning.} 
Figure~\ref{fig:supp_case_relative} illustrates relative-attribute reasoning. Here, the target is defined not by its standalone content, but by its relative ranking or temporal order among multiple candidate events. The model is required to comprehensively scan the video, compare measurable properties or track sequences such as identifying a comparative extremum or tracking the ordinal occurrence of an action, and then localize the correct instance.

\begin{figure*}[htbp]
  \centering
  \includegraphics[width=0.95\linewidth]{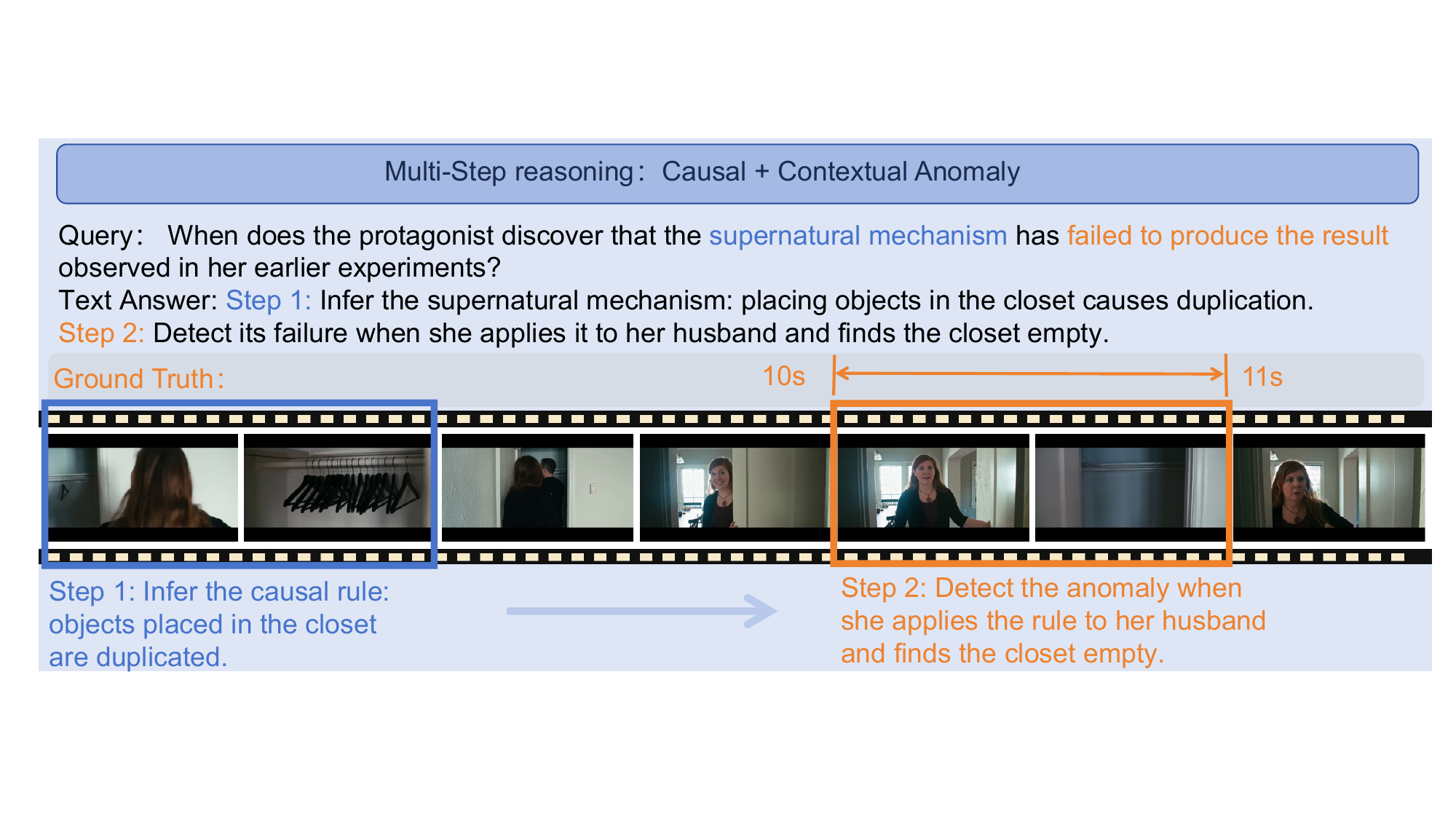}
  \vspace{-5pt}
  \caption{\textbf{An example of Multi-step reasoning.} The model first uses \textit{Causal} reasoning to infer the supernatural duplication mechanism, then detects a \textit{Contextual Anomaly} when the protagonist discovers that it has failed to produce the expected result for her husband. The target is the moment when she discovers the empty closet.}
  \label{fig:supp_case_multistep}
\end{figure*}

\noindent\textbf{Multi-Step Reasoning.} 
While the previous examples require a single atomic inference step, real-world user queries often involve compositional logic. Figure~\ref{fig:supp_case_multistep} demonstrates a multi-step reasoning case that combines causal reasoning with contextual anomaly detection. The model must first infer the supernatural mechanism from the protagonist's earlier experiments: placing objects in the closet causes them to duplicate. It must then use this causal rule to recognize the anomaly when the mechanism fails to produce the expected result for her husband, localizing the moment when she discovers the empty closet.

\FloatBarrier

\section{Evaluation Protocol}
\label{sec:supp_eval_details}

\subsection{Model Input and Sampling Settings}
\label{sec:supp_impl_details}


In this section, we detail the specific implementation settings for evaluating API-based and locally evaluated multimodal large language models (MLLMs) in our benchmark. We use model-specific official or default input protocols and common temporal-grounding metrics. The comparison characterizes complete systems under their supported settings; it does not isolate architecture or reasoning ability at an equal frame or token budget.

For API-based models, we prompt Gemini-3-Pro to output timestamps in the ``MM:SS'' format, following the best practices outlined in the official API documentation. The model samples videos at 1 FPS and uses the default \texttt{mediaResolution} parameter, which tokenizes each frame into 258 tokens. For GPT-5.5~\cite{openai2026gpt55}, we prepend textual timestamps such as ``Frame at 2.5s:'' to frames, with a baseline sampling rate of 1 FPS. Videos shorter than 50 seconds retain 1 FPS sampling; videos lasting 50--80 seconds are uniformly sampled to 50 frames; and longer videos retain 1 FPS sampling while every four consecutive frames are arranged into a $2\times2$ grid, capped at 50 composite images. We use temperature 0 and a maximum of 1,024 output tokens.

Qwen3.8-Omni-Flash and Seed2.0-Lite receive audio-visual inputs, whereas Qwen3.8-Flash and Seed2.0-Pro receive visual-only inputs. Each is evaluated under Direct, CoT, 2-step, and oracle-grounding prompts. Clock-format timestamps are normalized to seconds for scoring. Failed requests and invalid timestamp outputs are assigned zero.

Oracle grounding provides the ground-truth answer as the target description. Failed requests and invalid timestamp outputs are assigned zero.

For open-source Omni models, Qwen2.5-Omni, Qwen3-Omni, and HumanOmniV2 employ a dynamic sampling strategy: videos shorter than 384 seconds are sampled at 2 FPS, while videos longer than 384 seconds are fixed at a maximum of 768 frames. For these models, the minimum token count is uniformly set to 64, and the total token count is set to 14,336. Additionally, Ola-7B, ChronusOmni, and MiniCPM-O-2.6 uniformly sample a fixed 64 frames.

For open-source vision-language models, Qwen3-VL, Qwen2.5-VL, MiMo-VL-7B-RL, TimeLens-7B, and TimeLens-8B are sampled at 2 FPS with a maximum limit of 768 frames. Their minimum and total token counts are uniformly set to 64 and 14,336, respectively. Time-R1-7B uses Qwen2.5-VL-7B as the base model, samples at 2 FPS, and adaptively resizes the input resolution of each video to contain approximately 2.8 million pixels. Both TRACE and TimeSuite sample a fixed 128 frames, with their input resolutions uniformly set to 336 and 224, respectively. Video-R1-7B and VideoChat-R1, both based on Qwen2.5-VL, use a fixed 32 frames as video input.

\FloatBarrier

\section{Additional Results and Analyses}
\label{sec:extended_main_analyses}

This section presents the complete experimental results supporting the benchmark and model analyses in the main text, followed by additional self-verification and error-diagnosis experiments.

\subsection{Comparison with Descriptive Grounding Benchmarks}

\begin{table}[ht!]
\centering
\caption{Comparison of reported results on TimeLens-optimized benchmarks from~\cite{zhang2025timelens} with RVG-Bench. Gemini denotes Gemini-2.5-Pro for the first three benchmarks and Gemini-3-Pro for RVG-Bench. \textbf{Gap (pts)} reports the mIoU difference (Gemini minus TimeLens-8B) in percentage points. \textbf{Rel. gap} expresses this difference as a percentage of the TimeLens-8B score.}
\label{tab:saturation_comparison}
\small
\setlength{\tabcolsep}{2pt}
\begin{tabularx}{\columnwidth}{@{}>{\raggedright\arraybackslash}Xcccc@{}}
\toprule
\textbf{Benchmark} &
\textbf{Gemini} &
\textbf{TimeLens-8B} &
\textbf{Gap (pts)} &
\textbf{Rel.\ gap} \\
\midrule
Charades-TimeLens~\cite{zhang2025timelens}     & 52.8 & 55.2 & $-2.4$ & $-4.3\%$ \\
ActivityNet-TimeLens~\cite{zhang2025timelens}  & 58.1 & 53.2 & $+4.9$ & $+9.2\%$ \\
QVHighlights-TimeLens~\cite{zhang2025timelens} & 70.4 & 65.5 & $+4.9$ & $+7.5\%$ \\
\midrule
\textbf{RVG-Bench} & \textbf{44.4} & \textbf{30.1} & \textbf{+14.3} & \textbf{+47.5\%} \\
\bottomrule
\end{tabularx}
\end{table}

\FloatBarrier

\subsection{Modality Ablation}

\begin{table*}[t!]
\centering
\caption{Modality ablation results on the traditional TVG benchmark Charades-TimeLens and RVG-Bench. We report mIoU under the A+V, V-only, and A-only input settings. RVG-Bench results cover all single- and multi-step queries in the evaluation subset and are further divided into the Vis-Req, Aud-Req, and Both-Req subsets, along with the overall score.}
\label{tab:modality_ablation_comparison}
\small
\setlength{\tabcolsep}{4pt}
\resizebox{\textwidth}{!}{%
\begin{tabular}{ll c cccc}
\toprule
\multirow{3}{*}{\textbf{Model}} & \multirow{3}{*}{\textbf{Input}} & \textbf{Traditional} & \multicolumn{4}{c}{\textbf{RVG-Bench}} \\
\cmidrule(lr){3-3} \cmidrule(lr){4-7}
& & \textbf{Charades-TimeLens} & \multicolumn{3}{c}{\textbf{Subset by Required Modality}} & \multirow{2}{*}{\textbf{Overall}} \\
\cmidrule(lr){4-6}
& & \textbf{Overall} & \textbf{Vis-Req} & \textbf{Aud-Req} & \textbf{Both-Req} & \\
\midrule
\multirow{3}{*}{ChronusOmni}
& A+V    & 44.1 & 25.6 & 23.7 & 21.1 & \textbf{23.4} \\
& V-only & 41.5 & 23.8 & 18.2 & 18.3 & 20.8 \\
& A-only & 19.1 & 12.0 & 16.6 & 16.6 & 14.5 \\
\midrule
\multirow{3}{*}{Gemini-3-Pro}
& A+V    & 52.7 & 44.0 & 53.1 & 42.5 & \textbf{44.4} \\
& V-only & 48.2 & 46.2 & 24.9 & 33.7 & 38.4 \\
& A-only & 14.0 & 11.7 & 22.6 & 19.6 & 16.3 \\
\midrule
\multirow{3}{*}{Qwen3-Omni}
& A+V    & 37.2 & 8.8 & 15.9 & 11.2 & \textbf{10.7} \\
& V-only & 37.8 & 10.9 & 10.0 & 10.2 & 10.5 \\
& A-only & 19.3 & 4.4  & 13.5 & 8.1  & 7.0 \\
\bottomrule
\end{tabular}%
}
\end{table*}

\FloatBarrier

\subsection{Full Oracle-Grounding Results}
\label{sec:supp_full_oracle}

\begin{table*}[t]
\centering
\caption{Default and oracle-grounding mIoU. Ground subscripts under Overall denote gains over Default. The final row is the unweighted mean across the 22 listed models.}
\label{tab:full_default_vs_ground_miou}
\small
\setlength{\tabcolsep}{3pt}
\resizebox{\textwidth}{!}{%
\begin{tabular}{lrrrrrr}
\toprule
\multirow{2}{*}{\textbf{Model}} & \multicolumn{2}{c}{\textbf{S-Step}} & \multicolumn{2}{c}{\textbf{M-Step}} & \multicolumn{2}{c}{\textbf{Overall}} \\
\cmidrule(lr){2-3}\cmidrule(lr){4-5}\cmidrule(lr){6-7}
 & Default & Ground & Default & Ground & Default & Ground \\
\midrule
Qwen3.8-Omni-Flash & 51.3 & 60.7 & 40.6 & 50.8 & 49.6 & 59.1$_{+9.5}$ \\
Seed2.0-Lite & 49.0 & 59.8 & 36.9 & 46.3 & 47.0 & 57.6$_{+10.6}$ \\
Gemini-3-Pro & 46.1 & 65.7 & 35.8 & 54.7 & 44.4 & 63.9$_{+19.5}$ \\
Seed2.0-Pro & 42.6 & 57.7 & 38.0 & 51.0 & 41.9 & 56.6$_{+14.7}$ \\
GPT-5.5 & 39.5 & 56.4 & 32.8 & 40.5 & 38.5 & 53.6$_{+15.1}$ \\
Qwen3.8-Flash & 38.6 & 50.3 & 36.2 & 47.6 & 38.2 & 49.8$_{+11.6}$ \\
ChronusOmni & 25.0 & 37.5 & 15.7 & 23.4 & 23.4 & 35.2$_{+11.7}$ \\
Qwen3-Omni & 12.2 & 20.9 & 2.8 & 4.5 & 10.7 & 18.2$_{+7.5}$ \\
Qwen2.5-Omni & 5.8 & 10.1 & 1.4 & 3.3 & 5.1 & 9.0$_{+3.9}$ \\
HumanOmniV2 & 6.1 & 7.6 & 0.3 & 0.5 & 5.1 & 6.4$_{+1.3}$ \\
Ola-7B & 4.9 & 5.4 & 2.4 & 7.9 & 4.5 & 5.8$_{+1.4}$ \\
MiniCPM-O-2.6 & 2.0 & 4.4 & 3.3 & 3.9 & 2.2 & 4.3$_{+2.1}$ \\
TimeLens-8B & 31.9 & 46.7 & 21.3 & 39.0 & 30.1 & 45.4$_{+15.3}$ \\
Qwen3-VL-8B & 28.3 & 37.5 & 16.4 & 32.1 & 26.3 & 36.6$_{+10.3}$ \\
TimeLens-7B & 23.5 & 40.9 & 13.1 & 28.4 & 21.8 & 38.8$_{+17.0}$ \\
Time-R1-7B & 21.1 & 29.1 & 12.7 & 21.6 & 19.7 & 27.9$_{+8.2}$ \\
TRACE & 14.9 & 26.5 & 10.1 & 22.6 & 14.1 & 25.9$_{+11.8}$ \\
Qwen2.5-VL-7B & 15.6 & 15.9 & 2.3 & 2.0 & 13.4 & 13.6$_{+0.2}$ \\
VideoChat-R1 & 13.0 & 15.1 & 6.6 & 12.9 & 11.9 & 14.7$_{+2.8}$ \\
TimeSuite & 10.6 & 19.8 & 3.6 & 8.3 & 9.4 & 18.0$_{+8.5}$ \\
Video-R1-7B & 7.3 & 9.3 & 1.8 & 3.1 & 6.4 & 8.3$_{+1.9}$ \\
MiMo-VL-7B-RL & 2.0 & 1.8 & 0.2 & 1.6 & 1.7 & 1.7$_{0.0}$ \\
\midrule
\textbf{Avg. (22 models)} & 22.3 & 30.9 & 15.2 & 23.0 & 21.2 & 29.6$_{+8.4}$ \\
\bottomrule
\end{tabular}%
}
\end{table*}
\Cref{tab:full_default_vs_ground_miou} reports the full results corresponding to \cref{fig:oracle_grounding_trend}. Oracle target descriptions generally improve overall grounding performance, with the 22-model mean increasing from 21.2 to 29.6 mIoU. Gemini-3-Pro improves from 44.4 to 63.9, and TimeLens-8B from 30.1 to 45.4. However, models with weaker localization capabilities continue to struggle even when the correct target is supplied: Qwen2.5-Omni, HumanOmniV2, and MiniCPM-O-2.6 achieve only 9.0, 6.4, and 4.3 mIoU, respectively. These results indicate that making the target content explicit is beneficial, while accurate temporal localization remains a necessary capability for solving the task.

The 15-model mean in \cref{fig:oracle_grounding_trend} comprises Qwen3.8-Omni-Flash, Qwen3.8-Flash, Seed2.0-Lite, Seed2.0-Pro, Gemini-3-Pro, GPT-5.5, ChronusOmni, Qwen3-Omni, Qwen2.5-Omni, TimeLens-8B, Qwen3-VL-8B, Time-R1-7B, Qwen2.5-VL-7B, TimeSuite, and Video-R1-7B.

\FloatBarrier

\subsection{Full Task-Decomposition Results}
\label{sec:supp_full_tts}

\begin{table*}[t]
\centering
\caption{mIoU under Direct, CoT, and 2-step prompting on matched queries within each model. The final row is the unweighted mean across the 15 listed models.}
\label{tab:direct_cot_2step_miou_summary}
\small
\setlength{\tabcolsep}{3.2pt}
\resizebox{\textwidth}{!}{%
\begin{tabular}{lrrrrrrrrr}
\toprule
\multirow{2}{*}{\textbf{Model}} & \multicolumn{3}{c}{\textbf{S-Step}} & \multicolumn{3}{c}{\textbf{M-Step}} & \multicolumn{3}{c}{\textbf{Overall}} \\
\cmidrule(lr){2-4}\cmidrule(lr){5-7}\cmidrule(lr){8-10}
 & Direct & CoT & 2-step & Direct & CoT & 2-step & Direct & CoT & 2-step \\
\midrule
Qwen3.8-Omni-Flash & 48.8 & 51.3 & 46.0 & 37.4 & 40.6 & 37.4 & 47.0 & 49.6 & 44.6 \\
Seed2.0-Lite & 48.2 & 49.0 & 46.4 & 38.3 & 36.9 & 37.7 & 46.5 & 47.0 & 44.9 \\
Gemini-3-Pro & 44.1 & 46.1 & 51.1 & 39.0 & 35.8 & 43.2 & 43.2 & 44.4 & 49.8 \\
Seed2.0-Pro & 42.9 & 42.6 & 42.1 & 38.0 & 38.0 & 36.2 & 42.1 & 41.9 & 41.1 \\
GPT-5.5 & 38.6 & 39.5 & 38.9 & 36.8 & 32.8 & 36.4 & 38.4 & 38.5 & 38.6 \\
Qwen3.8-Flash & 40.1 & 38.6 & 35.7 & 35.2 & 36.2 & 37.6 & 39.3 & 38.2 & 36.0 \\
ChronusOmni & 22.3 & 25.0 & 23.4 & 13.1 & 15.7 & 16.9 & 20.8 & 23.4 & 22.4 \\
Qwen3-Omni & 15.4 & 12.2 & 15.1 & 2.4 & 2.8 & 3.0 & 13.2 & 10.7 & 13.1 \\
Qwen2.5-Omni & 6.8 & 5.8 & 8.5 & 1.8 & 1.4 & 2.0 & 6.0 & 5.1 & 7.4 \\
TimeLens-8B & 26.2 & 31.9 & 31.7 & 13.9 & 21.3 & 20.8 & 24.2 & 30.1 & 29.9 \\
Qwen3-VL-8B & 27.4 & 28.3 & 28.3 & 18.2 & 16.4 & 19.0 & 25.9 & 26.3 & 26.8 \\
Time-R1-7B & 21.0 & 21.1 & 21.1 & 13.1 & 12.7 & 12.5 & 19.7 & 19.7 & 19.7 \\
Qwen2.5-VL-7B & 14.7 & 15.6 & 14.4 & 1.5 & 2.3 & 2.7 & 12.5 & 13.4 & 12.4 \\
TimeSuite & 13.4 & 10.6 & 13.2 & 3.5 & 3.6 & 4.1 & 11.8 & 9.4 & 11.7 \\
Video-R1-7B & 7.3 & 7.3 & 8.0 & 0.1 & 1.8 & 1.4 & 6.1 & 6.4 & 7.0 \\
\midrule
\textbf{Avg. (15 models)} & 27.8 & 28.3 & 28.3 & 19.5 & 19.9 & 20.7 & 26.5 & 26.9 & 27.0 \\
\bottomrule
\end{tabular}%
}
\end{table*}
\Cref{tab:direct_cot_2step_miou_summary} separates the effects of prompting on single- and multi-step queries. Relative to CoT, 2-step improves multi-step mIoU for 10 of 15 models, compared with 6 of 15 on single-step queries. The corresponding mean changes are $+0.83$ and $-0.05$ points, producing a small overall gain of $0.09$ points. Relative to Direct, 2-step also improves multi-step mIoU for 10 of 15 models, by $1.22$ points on average. Some models exhibit opposing effects across subsets: ChronusOmni's single-step score decreases from 25.0 to 23.4 while its multi-step score increases from 15.7 to 16.9. Seed2.0-Lite similarly drops from 49.0 to 46.4 on single-step queries but improves from 36.9 to 37.7 on multi-step queries.

An improvement over CoT does not always translate into an improvement over Direct. Qwen3-Omni scores 13.2, 10.7, and 13.1 under Direct, CoT, and 2-step, respectively; TimeSuite scores 11.8, 9.4, and 11.7. For these models, decomposition largely recovers performance lost under CoT. By contrast, Gemini-3-Pro improves over both alternatives, from 43.2 with Direct and 44.4 with CoT to 49.8 with 2-step. TimeLens-8B benefits from CoT over Direct (24.2 to 30.1), with little further change under 2-step (29.9).

The oracle comparison provides a complementary perspective. Qwen3.8-Omni-Flash, Qwen3.8-Flash, Seed2.0-Lite, and Seed2.0-Pro all have lower overall mIoU under 2-step than CoT, while oracle target descriptions improve their scores by 9.5, 11.6, 10.6, and 14.7 points. These comparisons assess the combined process of generating and grounding an intermediate description, rather than answer correctness alone. The contrast highlights a gap between benefiting from a supplied target description and reliably producing and using one. It therefore motivates improving both evidence-grounded target inference and the alignment of intermediate descriptions with temporal spans, so that task decomposition can translate into more reliable grounding.

\FloatBarrier

\subsection{Recall at Stricter IoU Thresholds}
\label{sec:strict_recall}
\begin{table*}[t]
\centering
\caption{Default-CoT performance at progressively stricter temporal-overlap thresholds. R@0.3, R@0.5, and R@0.7 are percentages.}
\label{tab:strict_recall}
\small
\setlength{\tabcolsep}{7pt}
\resizebox{\textwidth}{!}{%
\begin{tabular}{lrrrr}
\toprule
Model & mIoU & R@0.3 & R@0.5 & R@0.7 \\
\midrule
Qwen3.8-Omni-Flash & 49.6 & 68.0 & 51.8 & 33.4 \\
Seed2.0-Lite & 47.0 & 62.6 & 49.2 & 30.5 \\
Gemini-3-Pro & 44.4 & 61.1 & 46.1 & 27.8 \\
Seed2.0-Pro & 41.9 & 58.2 & 44.8 & 27.1 \\
GPT-5.5 & 38.5 & 55.5 & 39.5 & 21.1 \\
Qwen3.8-Flash & 38.2 & 52.8 & 39.1 & 24.5 \\
ChronusOmni & 23.4 & 33.4 & 20.3 & 8.6 \\
Qwen3-Omni & 10.7 & 14.4 & 8.1 & 2.7 \\
Qwen2.5-Omni & 5.1 & 7.3 & 2.9 & 1.4 \\
HumanOmniV2 & 5.1 & 7.1 & 4.0 & 1.4 \\
Ola-7B & 4.5 & 6.1 & 2.7 & 1.1 \\
MiniCPM-O-2.6 & 2.2 & 2.7 & 1.6 & 0.6 \\
TimeLens-8B & 30.1 & 41.7 & 30.8 & 17.8 \\
Qwen3-VL-8B & 26.3 & 35.0 & 26.8 & 15.3 \\
TimeLens-7B & 21.8 & 31.1 & 19.8 & 10.6 \\
Time-R1-7B & 19.7 & 29.4 & 16.3 & 7.3 \\
TRACE & 14.1 & 20.8 & 11.0 & 4.9 \\
Qwen2.5-VL-7B & 13.4 & 20.4 & 11.3 & 4.9 \\
VideoChat-R1 & 11.9 & 17.4 & 9.4 & 3.1 \\
TimeSuite & 9.4 & 13.8 & 7.7 & 2.6 \\
Video-R1-7B & 6.4 & 9.3 & 4.3 & 1.3 \\
MiMo-VL-7B-RL & 1.7 & 2.4 & 0.9 & 0.3 \\
\bottomrule
\end{tabular}%
}
\end{table*}
Stricter overlap requirements lower recall across model groups. Even the leading Qwen3.8-Omni-Flash reaches 51.8\% at IoU $\geq 0.5$ and 33.4\% at IoU $\geq 0.7$, compared with 68.0\% at IoU $\geq 0.3$. Thus, relatively strong coarse localization does not imply consistently precise event boundaries.
\FloatBarrier

\subsection{Self-Verification}
\label{sec:detailed_test_time_scaling}

We further investigate whether additional test-time computation can improve temporal grounding through self-verification. Specifically, we adopt a best-of-$N$ setting in which each model independently generates $N$ candidate predictions, evaluates the candidates, and selects the one it considers most reliable. We evaluate $N\in\{1,3,5,10\}$ on the full evaluation set; $N=1$ corresponds to the standard single-generation CoT setting and matches the main results in \cref{tab:main_miou}.

\begin{figure}[t]
  \centering
  \includegraphics[width=0.78\linewidth]{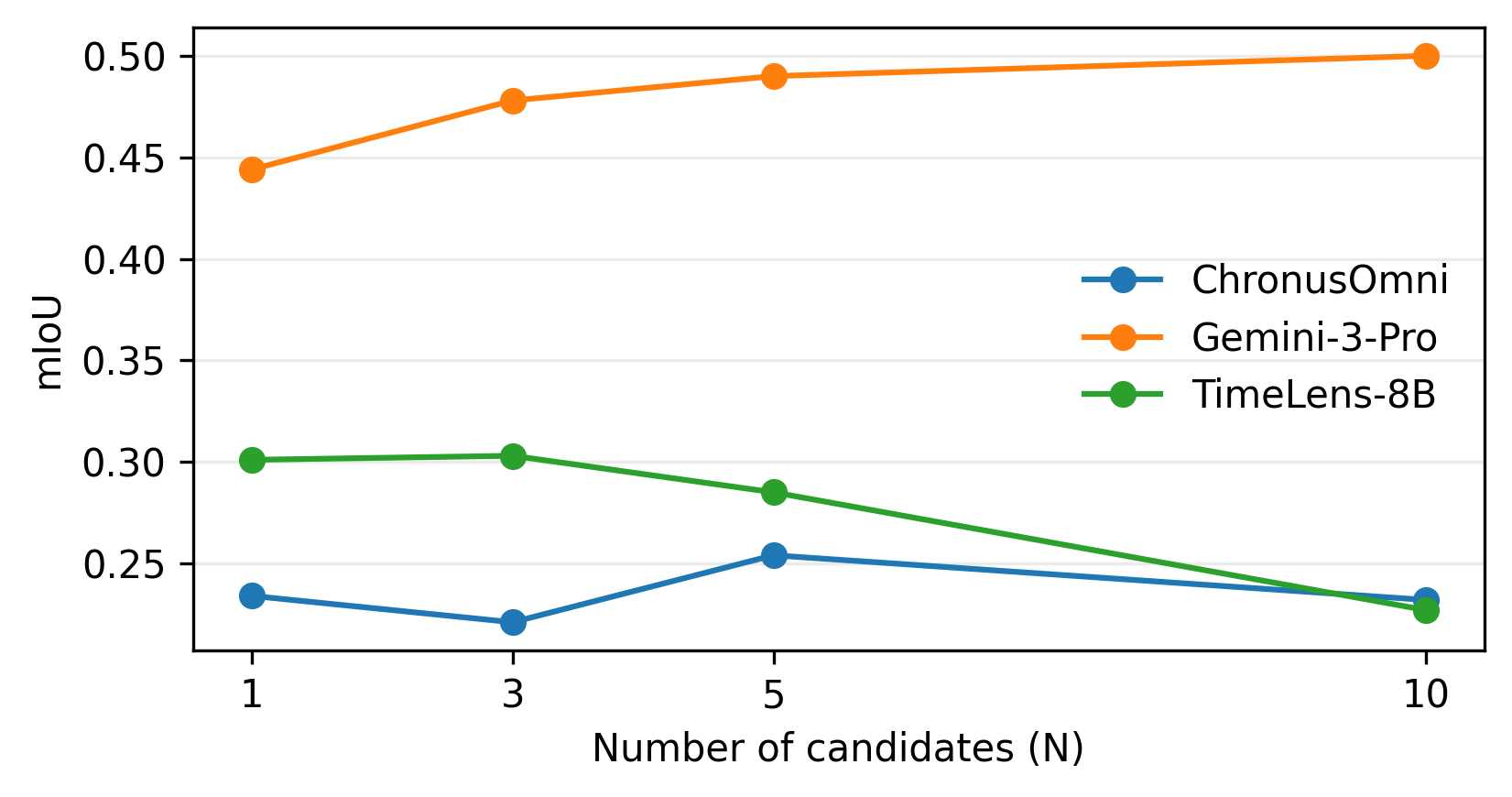}
  \caption{Best-of-$N$ self-verification on the full evaluation set, where $N$ is the number of candidate predictions per query. The $N=1$ scores match the CoT results in \cref{tab:main_miou}.}
  \label{fig:tts_miou}
\end{figure}

As shown in \cref{fig:tts_miou}, the effectiveness of self-verification varies substantially across models. Gemini-3-Pro improves consistently from 44.4 mIoU at $N=1$ to 47.8, 49.0, and 50.0 at $N=3,5,10$, respectively, for an overall gain of 5.6 points. This suggests that it possesses useful internal signals for comparing alternative reasoning paths and temporal boundaries. In contrast, the gains do not transfer reliably to the evaluated open-source models. TimeLens-8B increases slightly from 30.1 to 30.3 at $N=3$, then decreases to 28.5 at $N=5$ and 22.7 at $N=10$. ChronusOmni fluctuates from 23.4 to 22.1, 25.4, and 23.2 as $N$ increases, showing no consistent upward trend.

These results expose a generation--verification gap. Merely sampling more candidates is insufficient when a model cannot reliably distinguish a correct reasoning path and well-aligned temporal interval from plausible but erroneous alternatives. Self-verification therefore appears promising for sufficiently capable proprietary models. However, improving the verifier or the model's intrinsic ability to assess its own outputs remains necessary before additional test-time computation can robustly benefit current open-source systems.

\FloatBarrier

\subsection{Error Diagnosis}
\label{sec:supp_error_diagnosis}
\label{sec:judge_agreement}

\noindent\textbf{Protocol and evaluation subset.}
Aggregate grounding metrics do not indicate which stage of the reasoning--grounding pipeline causes a failure. We therefore conduct fine-grained error diagnosis on a subset of approximately 700 examples selected by stratified random sampling in proportion to the reasoning-category distribution. We use human-validated query--interval annotations and the corresponding model predictions, regarding predictions with IoU $\geq 0.5$ as successful localizations. We distinguish four error types: understanding errors indicate a failure to interpret the query intent; perception errors indicate missing or incorrect audio-visual evidence; reasoning errors indicate that the evidence is perceived but connected incorrectly; and localization errors indicate that the target content is identified but its temporal boundaries are inaccurate.

For diagnosis with Gemini-3-Pro~\cite{googledeepmind2026gemini3pro} as the judge, each input comprises the video captions, query, ground-truth textual answer, and model outputs, predicted timestamps, and precomputed IoUs from default CoT, oracle grounding, and two-step prediction. The default-CoT failure analysis uses evidence from all three settings. For unsuccessful predictions with valid diagnostic evidence from all three settings, we report the earliest failed stage under the Understanding $\prec$ Perception $\prec$ Reasoning $\prec$ Localization hierarchy; error-type proportions are computed among these failed localizations.

For Qwen3.8-Omni-Flash, Qwen3.8-Flash, Seed2.0-Lite, and Seed2.0-Pro, GPT-5.5 diagnoses the default-CoT responses using the video captions, reference annotations, and model outputs for this setting. We report each error label's frequency among valid judgments, including successful and unsuccessful localizations; labels may overlap.

\noindent\textbf{Model-specific failure patterns.}
Perception remains a substantial obstacle for several evaluated systems. In the default-CoT failure analysis with Gemini-3-Pro as the judge, perception errors account for 44.9\% of ChronusOmni's failures, 45.9\% of Qwen3-Omni's failures, 53.0\% of Qwen2.5-Omni's failures, and 79.9\% of Video-R1-7B's failures. These results highlight difficulties in extracting subtle and temporally dispersed evidence, including in systems with access to both audio and video.

The GPT-5.5 diagnoses also reveal prominent localization difficulties across the Qwen and Seed models (Table~\ref{tab:new_models_diagnosis}). Localization is the most frequent error label for Qwen3.8-Omni-Flash, Qwen3.8-Flash, and Seed2.0-Lite, appearing in 34.6\%, 33.2\%, and 38.6\% of valid judgments, respectively. Seed2.0-Pro exhibits comparable frequencies of localization and reasoning errors (29.0\% and 28.9\%). Together with the gains and remaining errors under oracle grounding, these findings suggest that improving target inference alone is insufficient: models must also accurately align the inferred target with its temporal extent.

\begin{table}[t]
\caption{Error-label frequencies (\%) assigned by GPT-5.5 to default-CoT responses. U/P/R/L denote understanding, perception, reasoning, and localization.}
\label{tab:new_models_diagnosis}
\centering
\small
\setlength{\tabcolsep}{4pt}
\begin{tabular}{lrrrr}
\toprule
Evaluated model & U & P & R & L \\
\midrule
Qwen3.8-Omni-Flash & 3.2 & 13.8 & 16.9 & 34.6 \\
Qwen3.8-Flash & 2.2 & 20.9 & 24.9 & 33.2 \\
Seed2.0-Lite & 7.0 & 16.7 & 22.5 & 38.6 \\
Seed2.0-Pro & 5.9 & 15.5 & 28.9 & 29.0 \\
\bottomrule
\end{tabular}
\end{table}

\noindent\textbf{Agreement across judges.}
To assess the consistency of the identified error patterns, we compare Gemini-3-Pro preview and GPT-5.5 (\texttt{gpt-5.5-2026-04-24}) diagnoses for Gemini-3-Pro and Qwen3-VL-8B across default CoT, oracle grounding, and two-step prediction. Both judges use video captions, reference annotations, and model outputs.
For each matched query and setting, the primary diagnostic label is the first reported error type, or \textit{no error} if no error is reported. Agreement is computed over these five labels, excluding failed or invalid judge outputs; the four error-category frequencies are pooled across the three settings.

Primary-label agreement is 80.11\% for Gemini-3-Pro and 70.03\% for Qwen3-VL-8B, with Cohen's $\kappa$ of 0.704 and 0.600, respectively (Table~\ref{tab:judge_agreement}).
For both evaluated models, both judges give the same full error-category ranking: localization $>$ perception $>$ reasoning $>$ understanding. Localization is therefore the most frequent primary error in this cross-setting analysis, although localization and perception remain close for Qwen3-VL-8B, particularly under GPT-5.5.
The consistent rankings support temporal alignment as a recurring difficulty across the two models and judges, complementing the remaining localization errors under oracle target descriptions.

\begin{table}[t]
\caption{Agreement between Gemini-3-Pro and GPT-5.5 on primary diagnostic labels across CoT, oracle grounding, and 2-step prediction. Labels comprise four error types and \textit{no error}; $\kappa$ is Cohen's kappa.}
\label{tab:judge_agreement}
\centering
\small
\setlength{\tabcolsep}{4pt}
\begin{tabular}{lrr}
\toprule
Evaluated model & Agreement (\%) & $\kappa$ \\
\midrule
Gemini-3-Pro & 80.11 & 0.704 \\
Qwen3-VL-8B & 70.03 & 0.600 \\
\bottomrule
\end{tabular}
\end{table}

\noindent\textbf{Reasoning in successful localizations.}
For successful predictions (IoU $\geq 0.5$), the Gemini-3-Pro judge distinguishes \textit{right-reasoning} (RR), where the localization is supported by a correct reasoning path, from \textit{wrong-reasoning} (RW), where the predicted interval is correct despite a flawed intermediate rationale. This distinction helps identify successes that may arise from shortcuts or spurious correlations rather than reliable reasoning. Among successful localizations, Qwen2.5-Omni and TimeSuite receive RW labels for 25.0\% and 27.5\% of predictions, respectively. These labels assess the models' generated explanations and are separate from the primary diagnostic labels used in the cross-judge comparison.

\FloatBarrier

\section{Prompt Templates}
\label{sec:supp_prompt_templates}

This section presents evaluation and construction prompt templates for RVG-Bench. For readability, some templates are abridged and summarize the core instructions rather than reproduce the complete experimental prompts verbatim. We will release the full, unabridged prompts for all construction and evaluation stages, together with code for data construction and evaluation.

\subsection{Evaluation Prompts}
\label{sec:supp_eval_prompts}

To comprehensively evaluate the models' reasoning and localization capabilities, we design four distinct types of evaluation prompts: Default Chain-of-Thought (CoT), Direct, Two-step, and Ground (Oracle Answer). For the detailed definitions of the different evaluation prompts, please refer to the Experiments section in the main manuscript. The templates below illustrate these four evaluation settings.

\begin{promptbox}{Default (CoT) Prompt}
\begin{lstlisting}[style=promptstyle]
You are an AI assistant for audio-visual video understanding.
Your task is to identify the specific timestamp interval in the video that serves as the direct answer to the user's question.
Task Definition:
1. Consider all available information sources, including visual actions, dialogue, sound effects, and background audio.
2. Locate the precise segment that constitutes the answer to the user's query.
3. Provide the result as a single JSON object with the following fields:
   3.1. "reasoning": Briefly explain what content is in the selected segment and why it is the answer.
   3.2. "timestamp": Start time and end time in seconds from the beginning of the video, formatted as 12.00-25.50.
Question: {question}
Now, based on the full video content, identify the time span that contains the answer to the question.
\end{lstlisting}
\end{promptbox}

\begin{promptbox}{Direct Prompt}
\begin{lstlisting}[style=promptstyle]
You are an AI assistant for audio-visual video understanding.
Your task is to identify the specific timestamp interval in the video that serves as the direct answer to the user's question.
Task Definition:
1. Consider all available information sources, including visual actions, dialogue, sound effects, and background audio.
2. Locate the precise segment that constitutes the answer to the user's query.
3. Provide the result as a single timestamp interval in seconds from the beginning of the video.
   3.1. Output only the interval, with no extra text.
   3.2. Use the format start-end, for example, 12.00-25.50.
Question: {question}
Now, based on the full video content, identify the time span that contains the answer to the question.
\end{lstlisting}
\end{promptbox}

\begin{promptbox}{2-Step Prompt: Step 1 (Answer Generation)}
\begin{lstlisting}[style=promptstyle]
You are an AI assistant for audio-visual video understanding.
The user will ask a question about the video.
The answer to this QA is not abstract; it must be grounded in a specific video segment that serves as the direct answer.
Your goal in this step is to reason about what the correct answer refers to in the video and describe the specific video moment or segment that directly serves as the answer.
Instruction:
1. Do not provide any timestamps or time ranges in any format.
2. Output only a single JSON object with the following fields:
   2.1. "reason": A brief explanation of why this video moment or segment is the answer.
   2.2. "answer_desc": A description of the specific video moment or segment that directly serves as the answer.
Question: {question}
Now, based on the full video content, describe the specific video moment or segment that directly serves as the answer.
\end{lstlisting}
\end{promptbox}

\begin{promptbox}{2-Step Prompt: Step 2 (Temporal Localization)}
\begin{lstlisting}[style=promptstyle]
You are an expert in video temporal grounding.
The user will ask a question about the video.
The answer to this QA is not abstract; it must be grounded in a specific video segment that serves as the direct answer.
You are given the following information:
1. The video.
2. The user's question.
3. Prior-step analysis, including:
   3.1. Reason: why the answer should correspond to a particular moment or segment.
   3.2. Target Description: what to look for in the video as the answer to the user's question.
Your task is to use both the Reason and the Target Description as guidance to find the exact start and end timestamps of the segment that directly answers the question.
Output Requirement:
1. Output only a single JSON object with the following fields:
   1.1. "reasoning": A brief grounding justification.
   1.2. "timestamp": Start and end time in seconds from the beginning of the video, formatted as "12.00-25.50".
Question: {question}
Answer: {answer}
Now locate the precise time interval where the Answer content occurs in the video.
\end{lstlisting}
\end{promptbox}

\begin{promptbox}{Ground (Oracle Answer) Prompt}
\begin{lstlisting}[style=promptstyle]
You are an expert in temporal grounding for videos.
You will be provided with a Question and its correct Answer.
Your sole task is to locate the video segment that corresponds to the provided Answer.
Core Instruction:
1. The content to locate is already explicitly described in the provided Answer text.
2. Do not re-analyze the Question to derive a new answer.
3. Treat the Answer as a visual or audio description and find the precise timestamp interval where this description occurs in the video.
Output Requirement:
1. Output only a single JSON object with the following field:
   1.1. "timestamp": Start and end time in seconds from the beginning of the video, formatted as "12.00-25.50".
Question: {question}
Answer: {answer}
Now locate the precise time interval where the Answer content occurs in the video.
\end{lstlisting}
\end{promptbox}

We apply model-specific prompt adaptations to ensure a fair evaluation aligned with each model's architectural constraints and official training formats:
\begin{itemize}
    \item \textbf{Vision-Language (VL) Models:} For models lacking audio comprehension, we modify the ``Multimodal Analysis'' instruction to exclude references to dialogue, sound effects, and background audio, restricting the requirement to visual cues only.
    \item \textbf{Gemini Models:} We adjust the output format requirement to predict timestamps in the ``MM:SS'' format instead of absolute seconds, strictly adhering to its official API best practices.
    \item \textbf{Reasoning Models (Time-R1-7B, Video-R1-7B, VideoChat-R1):} Following their official repository guidelines, we explicitly constrain the output structure by requiring the models to wrap their reasoning and final predictions within \texttt{<think>} and \texttt{<answer>} tags.
\end{itemize}

\FloatBarrier

\subsection{Construction Prompts}
\label{sec:supp_construction_details}

This section provides prompt details for the automated construction pipeline of RVG-Bench, complementing the overview in Section~\ref{sec:benchmark-construction} of the main manuscript. The prompts orchestrate Gemini-2.5-Pro and Gemini-3-Pro across the generation and refinement stages. For presentation clarity, long prompts have been abridged to summarize their core instructions and intended behavior; the versions shown here do not reproduce every detail of the full prompts used in data construction. The full human-validation protocol and annotation interface are presented earlier in Appendix~\ref{sec:supp_human_validation}.

\subsubsection{Dataset Construction Prompts}
\label{sec:supp_dataset_prompts}

As described in the main text, our data synthesis pipeline relies on carefully engineered prompts to elicit deep narrative understanding and reasoning capabilities from the models. We present the corresponding prompt templates below, abridged where needed for readability.

\paragraph{Stage 1: Multi-view Caption Generation.}
To reduce the cognitive load on the query generator, we first parse the raw video into timestamped dense captions covering visual, audio, and joint events.
\begin{promptbox}{Multi-view Caption Generation Prompt}
\begin{lstlisting}[style=promptstyle]
You are an advanced video analysis system.
Your task is to perform a complete and seamless multi-view analysis by partitioning the entire video into continuous event segments.
Core Rules:
1. All timestamps must use the format "MM:SS".
2. Each event list, including visual_events, audio_events, and audiovisual_events, must continuously cover the video from 00:00 to the end without any gaps.
3. The end_time of one event must match the start_time of the next event within the same list.
4. Determine semantic boundaries autonomously.
5. The three event lists may have independent cut points based on the semantic boundaries of their own modalities.
Multimodal Synthesis:
1. For audiovisual_events, do not simply concatenate visual and audio descriptions.
2. Analyze the combined stream holistically.
3. Explicitly describe cross-modal relationships, such as causal links or reactions to sudden sounds.
Output Format:
1. Output only a JSON object that strictly follows the structure below.
{
  "visual_events": [
    {
      "start_time": "MM:SS",
      "end_time": "MM:SS",
      "description": "Concise, objective description of the visual content."
    }
  ],
  "audio_events": [
    {
      "start_time": "MM:SS",
      "end_time": "MM:SS",
      "description": "Description of sounds, such as sound effects or background music.",
      "transcription": "Exact speech if intelligible, '[unintelligible speech]' if speech exists but is unclear, or null if there is no speech."
    }
  ],
  "audiovisual_events": [
    {
      "start_time": "MM:SS",
      "end_time": "MM:SS",
      "description": "Holistic description of the combined audiovisual event, capturing meaningful interactions and dependencies."
    }
  ]
}
\end{lstlisting}
\end{promptbox}

\paragraph{Stage 1: Single-Step Query Generation.}
Using the generated captions as context, the model proposes candidate single-step query-interval pairs across various reasoning subcategories.

\begin{promptbox}{Single-step Query Generation Prompt}
\begin{lstlisting}[style=promptstyle]
You are an expert video reasoning analyst and benchmark designer.
Your task is to generate high-quality question-answer pairs based on a list of video events and a specific benchmark taxonomy.
Inputs:
1. Benchmark Taxonomy: {BENCHMARK_TAXONOMY}
2. Category Quality Requirements: {QUALITY_REQUIREMENTS}
3. Annotated Video Events: {events_json}
Generation Instructions:
1. Generate 1 to 3 high-quality QA pairs targeting specific L2 subcategories, such as "1.1 Causal" or "2.1 Character-centric".
2. Prioritize quality and diversity over quantity.
3. Do not force generation if the events do not support deep reasoning.
4. Ensure that each question matches the data access pattern of its chosen category.
   4.1. Relational questions should navigate dependencies between an Anchor and a Target.
   4.2. Intrinsic questions should infer complex, locally verifiable properties, such as emotion or symbolism.
   4.3. Relative questions should compare or sort a collection of events to identify an extremum or a specific order.
5. Each question must require inference rather than direct matching.
6. Avoid direct visual or audio descriptions of the target segment.
7. Each question must unambiguously resolve to a single temporal segment.
8. The answer must be verifiable from the provided context.
Output Format:
1. Return only a JSON list with the following schema.
2. The fields justification and quality_analyze are mandatory.
[
  {
    "chosen_category": "string (Full L2 category ID and name, for example, '1.3. Referential Reasoning')",
    "question": "string (Generated inferential question)",
    "answer": "string (Concise, newly generated description of the target event)",
    "time": ["MM:SS", "MM:SS"],
    "justification": "string (Brief explanation of why the QA pair fits the chosen category)",
    "quality_analyze": "string (Point-by-point verification of satisfied quality standards and avoided low-quality traps)"
  }
]
\end{lstlisting}
\end{promptbox}

\paragraph{Stage 1: Event Graph and Multi-Step Query Generation.}
Directly prompting for multi-hop reasoning often yields superficial results. Therefore, we introduce an intermediate \textit{event graph} representation to explicitly model inter-event dependencies before generating multi-step queries.

\begin{promptbox}{Event Graph Parsing Prompt}
\begin{lstlisting}[style=promptstyle]
You are an expert video narrative analyst.
Your task is to fuse raw multi-view video captions with high-level plot summaries to build a precise logic reasoning graph.
Your objective is to explain not only what happened at each timestamp, but also why it matters and who is involved.
Processing Workflow:
1. Read {key_info} first to understand the story arc.
2. Resolve pronouns and synthesize the visual and audio streams.
3. Group captions into logical event segments, each with a maximum length of 40 seconds.
4. Rewrite technical jargon into natural human storytelling.
5. Classify the narrative function of each segment, such as Setup, Climax, or Turning_Point.
6. Identify internal states, including emotion and intent.
7. Estimate physical metrics, including visual intensity and audio intensity on a 1 to 5 scale.
8. Tag anomalies and symbolic events.
9. Categorize connections between segments into relations such as Causal, Setup_Payoff, Reference, Contrast, or Similarity_Echo.
Inputs:
1. Captions: {caption}
2. Key Info: {key_info}
Output Format:
1. Output only a JSON object with the following schema.
{
  "segments": [
    {
      "id": "S1",
      "timestamp": ["MM:SS", "MM:SS"],
      "natural_summary": "Simple, active-voice sentence describing the event.",
      "narrative_function": "Setup | Climax | Turning_Point | Mistake | Realization | Conflict",
      "internal_states": {
        /* Fields for emotion_label, emotion_intensity, intent, and knowledge_state omitted for brevity */
      },
      "metrics": {
        /* Fields for visual intensity, audio intensity, and narrative importance omitted for brevity */
      },
      "tags": {
        /* Boolean flags such as is_anomaly and is_symbolic omitted for brevity */
      }
    }
  ],
  "logic_relations": [
    {
      "type": "Causal | Setup_Payoff | Reference | Contrast | Similarity_Echo",
      "source": "S_ID",
      "target": "S_ID",
      "description": "Natural sentence explaining the link.",
      "differentiation_note": "For Similarity_Echo or Contrast, explicitly explain the difference."
    }
  ]
}
\end{lstlisting}
\end{promptbox}

\begin{promptbox}{Multi-step Query Generation Prompt}
\begin{lstlisting}[style=promptstyle]
You are an expert video reasoning analyst.
Your task is to create temporal grounding questions that are logically complex to solve but linguistically simple to read.
Style Guidelines:
1. Use natural language and avoid academic jargon.
2. Prefer concrete event descriptions over abstract concepts.
3. Use the shortest possible anchor phrase that still remains unique.
Reasoning Requirements:
1. Every question must require two or more reasoning steps.
2. Construct questions by combining strategies from the predefined reasoning pools.
   2.1. Base Types: Causal, Sequential, Referential, Character-Centric, Anomaly, Symbolic, Comparative, Ordinal.
   2.2. Graph Types: Contrast, Setup-Payoff.
Question Construction Rules:
1. Each question must decompose into independent phrases that map to the reasoning types above.
2. The ground-truth segment must be the unique intersection of all phrases.
3. For every phrase, apply a necessity test by simulating its removal.
4. If removing a phrase causes ambiguity or shifts the target to a different segment, the phrase is valid.
5. If the ground-truth segment remains uniquely identifiable without a phrase, then that phrase is redundant and must be removed.
Examples:
1. Causal + Ordinal: "What caused the second argument?"
2. Character + Comparative + Causal: "Find the event that triggered the protagonist's most violent outburst."
Inputs:
1. VIDEO_ID: {video_id}
2. LOGIC_GRAPH: {logic_graph}
Output Format:
1. Output only a strict JSON object with the following schema.
2. Detailed self-reflection and necessity verification fields are required for each hop.
{
  "video_id": "string",
  "qa_items": [
    {
      "question": "string (Natural, short, and action-oriented)",
      "answer": "string (Evidence-based description)",
      "time": ["MM:SS", "MM:SS"],
      "hop_count": 2,
      "hop_steps": [
        {
          "step": 1,
          "reasoning_type": "string (for example, causal or ordinal)",
          "question_phrase": "string",
          "operation": "Select | Infer | Traverse",
          "from": "string",
          "to": "string",
          "why_needed": "string",
          "self_check_no_shortcut": {
            /* Fields for target direct description, semantic gap explanation, and risk level omitted for brevity */
          },
          "self_check_necessity_test": {
            /* Fields for single-phrase lock check and phrase ablation effect omitted for brevity */
          }
        }
      ]
    }
  ]
}
\end{lstlisting}
\end{promptbox}

\paragraph{Stage 2: Quality Refinement and Verification.}
To ensure the rigor of RVG-Bench, candidate queries undergo strict verification against four criteria: answerability, uniqueness, logical soundness, and absence of information leakage. We also employ specific prompts to refine the temporal boundaries and verify modality reliance.

\begin{promptbox}{Quality Verification Prompt}
\begin{lstlisting}[style=promptstyle]
You are a video QA quality assurance expert.
Your task is to verify the quality of a single question-answer pair and determine whether it is qualified.
Inputs:
1. Current Category: {category}
2. Timestamped Dense Captions: {captions}
3. QA Sample:
   3.1. question: {question}
   3.2. answer: {answer}
   3.3. time (GT): {time}
   3.4. justification: {justification}
Verification Criteria:
1. Answerability:
   Determine whether the question can be answered from the provided evidence and whether the answer refers to a concrete and temporally localizable event.
2. Uniqueness:
   Determine whether exactly one clear and reasonable segment in the video answers the question.
3. Logical Soundness:
   Determine whether the QA pair is factually consistent with the evidence, follows a valid reasoning path, and satisfies the category-specific requirements below.
   {category_requirement}
4. No Information Leakage:
   Determine whether the question avoids directly describing the visual or audio content of the target segment, so that the target must be inferred through reasoning.
Output Rules:
1. Output only valid JSON.
2. Do not output markdown.
3. Do not output extra keys.
4. Each of the four verification fields must be a single string in one of the following formats:
   4.1. "true; <reason>"
   4.2. "false; <reason>"
5. If any of the four fields is false, set is_valid to false.
6. If all four fields are true, set is_valid to true.
Output Format:
{
  "is_valid": true/false,
  "answerability": "true/false; reason",
  "uniqueness": "true/false; reason",
  "logical_soundness": "true/false; reason",
  "no_information_leakage": "true/false; reason"
}
\end{lstlisting}
\end{promptbox}

\begin{promptbox}{Timestamp Refinement Prompt}
\begin{lstlisting}[style=promptstyle]
You are a strict temporal grounding annotator for video QA.
You will be given the full video as input, including both audio and visual information, together with the following inputs.
Inputs:
1. Question
2. Answer, which defines the target event or state
3. Coarse Ground-Truth Time, which provides a rough time window from a coarse event graph
Task:
1. Refine the coarse ground-truth time into a precise ground-truth interval in the format ["MM:SS", "MM:SS"].
2. The refined interval must be the minimal continuous span that is both necessary and sufficient for the answer.
Rules:
1. Use the video content as the only source of truth.
2. You may shrink the interval within the coarse window.
3. You may shift slightly outside the coarse window if needed.
4. Choose boundaries as follows:
   4.1. The start time is the first clear moment at which the answer-defining evidence begins.
   4.2. The end time is the moment at which the answer-defining evidence ends or completes.
   4.3. Keep the interval as tight as possible, and avoid unnecessary lead-in or aftermath unless needed.
Output Rules:
1. Output only strict JSON.
Output Format:
{
  "status": "OK | NEED_MORE_CONTEXT",
  "new_time": ["MM:SS", "MM:SS"] | null,
  "change_type": "SHRINK_WITHIN | SHIFT_EARLIER | SHIFT_LATER | EXPAND | UNCHANGED | UNKNOWN",
  "why": "1-2 sentences describing the boundary choice",
  "confidence": "HIGH | MED | LOW"
}
Now process the following inputs:
1. QUESTION: {question}
2. ANSWER: {answer}
3. COARSE_GT_TIME: {coarse_time}
\end{lstlisting}
\end{promptbox}

\begin{promptbox}{Modality Verification Prompt}
\begin{lstlisting}[style=promptstyle]
You are an expert in video content analysis.
Your task is to analyze each QA pair and determine which modality or modalities are required to answer the question correctly based on the given time segment.
Inputs:
1. Caption Information:
   {caption_str}
2. QA List:
   {qa_list_str}
Analysis Goal:
For each QA pair, determine what information is required in order to:
1. Understand and answer the question.
2. Identify the specific time segment mentioned in the answer.
Modality Categories:
1. "only visual": Only visual information is required, such as analyzing what is shown on screen, reading text, or identifying objects or people.
2. "only audio": Only audio information is required, such as understanding speech content, identifying sounds, or recognizing music.
3. "both": Both visual and audio information are required, such as matching speech to visual content or understanding audiovisual synchronization.
Output Rules:
1. Return only a valid JSON array.
2. The array must contain one object per QA pair.
3. Preserve the same order as the input.
4. Each object must contain the following fields:
   4.1. "qa_index": the index number, such as 0, 1, 2, and so on.
   4.2. "required_modality": one of "only visual", "only audio", or "both".
   4.3. "reasoning": a brief explanation in one or two sentences of why this modality is required.
Now analyze the QA pairs and return your response.
\end{lstlisting}
\end{promptbox}

\FloatBarrier

\end{document}